\documentclass{article}

\PassOptionsToPackage{numbers, compress}{natbib}

\usepackage[preprint]{neurips_2026}

\usepackage[utf8]{inputenc} % allow utf-8 input
\usepackage[T1]{fontenc}    % use 8-bit T1 fonts
\usepackage{hyperref}       % hyperlinks
\usepackage{url}            % simple URL typesetting
\usepackage{booktabs}       % professional-quality tables
\usepackage{amsfonts}       % blackboard math symbols
\usepackage{nicefrac}       % compact symbols for 1/2, etc.
\usepackage{microtype}      % microtypography
\usepackage[table,xcdraw]{xcolor} % colors

\usepackage{multirow}

\usepackage{graphicx}
\usepackage{epsfig}
\usepackage{colortbl} % For table cell background colors
\usepackage{etoolbox} 
\usepackage{array}

\usepackage{pifont}
\usepackage{amsmath}
\usepackage{amssymb}
\usepackage{amsthm}
\usepackage{float} 
\usepackage[english]{babel}

\newcommand{\cmark}{\textcolor{green!60!black}{\ding{51}}}
\newcommand{\xmark}{\textcolor{red}{\ding{55}}}
\newcounter{algorithm}
\renewcommand{\thealgorithm}{\arabic{algorithm}}
\usepackage{amsmath,amsfonts,bm,amsthm}

\def\eqref#1{equation~\ref{#1}}
\def\1{\bm{1}}

\DeclareMathAlphabet{\mathsfit}{\encodingdefault}{\sfdefault}{m}{sl}
\SetMathAlphabet{\mathsfit}{bold}{\encodingdefault}{\sfdefault}{bx}{n}

\usepackage{amsmath}
\usepackage{amsfonts}
\usepackage{amssymb}

\newtheorem{theorem}{Theorem}

\newtheorem{lemma}{Lemma}
\newtheorem{proposition}{Proposition}

\title{Preference-Based Self-Distillation: Beyond KL Matching via Reward Regularization}

\author{%
  \textbf{Xin Yu}$^{1,2}$ \hspace{1.5em}
  \textbf{Liuchen Liao}$^{2}$ \hspace{1.5em}
  \textbf{Yiwen Zhang}$^{2}$\\[0.35em]
  \textbf{Yingchen Yu}$^{2}$ \hspace{1.5em}
  \textbf{Lingzhou Xue}$^{1}$\thanks{Corresponding authors. Emails: \texttt{lingzhou@psu.edu}, \texttt{guoqinzhen@bytedance.com}.\\ This work was completed while Xin Yu was an intern at TikTok.} \hspace{1.5em}
  \textbf{Qinzhen Guo}$^{2}$\footnotemark[1]\\[0.6em]
  $^{1}$The Pennsylvania State University\\
  $^{2}$TikTok
}

\begin{document}

\maketitle

\begin{abstract}
On-policy distillation is an efficient alternative to reinforcement learning, offering dense token-level training signals. However, its reliance on a stronger external teacher has driven recent work on on-policy self-distillation, where the same model serves as both teacher and student under different prompt contexts. Yet, existing self-distillation methods largely reduce learning to KL matching toward the context-augmented teacher model. This approach often suffers from training instability and can degrade reasoning performance over time. Moreover, self-distillation from the same model with prompt augmentation lacks the exploratory diversity provided by a genuine external teacher. To address these limitations, we move beyond fixed-teacher KL matching and propose \textbf{P}reference-\textbf{B}ased \textbf{S}elf-\textbf{D}istillation (\textbf{PBSD}), which revisits on-policy self-distillation through a reward-regularized perspective. Instead of directly matching the teacher distribution, we derive a reward-regularized objective whose analytic optimum is a reward-reweighted teacher distribution, yielding a target policy that improves over the original teacher under this objective. Practically, PBSD optimizes preference gaps between teacher and student samples while maintaining on-policy student sampling. We support this framework with a local statistical analysis of the induced preference-learning problem, clarifying when contextual self-distillation can provide more favorable comparison geometry than external-teacher distillation in our setting. Experiments on mathematical reasoning and tool-use benchmarks across multiple model scales demonstrate that PBSD consistently achieves the strongest average performance among comparable baselines, showing improved training stability over prior self-distillation baselines while preserving token efficiency. Code is available at \href{https://github.com/LucasXinYu/PBSD}{github.com/LucasXinYu/PBSD}.
\end{abstract}

\section{Introduction}
\label{sec:introduction}

On-policy distillation (OPD)~\citep{agarwal2024opd,lu2025onpolicydistillation} leverages a stronger teacher model to provide dense token-level learning signals along a student's sampled trajectories. It has emerged as an important paradigm for post-training large language models (LLMs)~\citep{song2026survey}, offering an efficient, structured alternative to reinforcement learning (RL) based optimization~\citep{lu2025onpolicydistillation,li2026sample}. Compared to RL-based post-training methods such as GRPO \citep{shao2024deepseekmath}, OPD is typically more token-efficient because it avoids repeated group rollouts and reward evaluations~\citep{lu2025onpolicydistillation,song2026survey}. However, standard OPD relies on a separate, typically larger teacher model and assumes a shared token vocabulary between teacher and student, which bottlenecks its computational efficiency and practical scalability~\citep{agarwal2024opd,fu2026revisiting,li2026rethinkingopd}. To overcome these limitations, on-policy self-distillation~\citep{zhao2026self,sang2026policy} has emerged as a compelling solution. Rather than querying a stronger external teacher model, on-policy self-distillation unifies the teacher and student models into a single architecture, with the teacher instantiated by conditioning the model on additional context \(\boldsymbol{c}\) in the prompt~\citep{zhao2026self}. Similar to OPD, existing on-policy self-distillation methods optimize the divergence between teacher and student distributions, often via a forward-KL objective~\citep{zhao2026self} or a reverse-KL objective~\citep{hubotter2026reinforcement,yang2026self}. More broadly, scalable self-distillation may help democratize capable open-source agents by reducing the reliance on expensive proprietary teachers or dense human supervision, unlocking new applications in multi-turn agent training~\citep{wang2026skillsd}, tool-use and conversational agents~\citep{wang2026openclaw}, and autonomous decision-making~\citep{afsharrad2026policy}.

\begin{figure}[t]
    \centering
    \includegraphics[width=0.69\linewidth,height=0.50\linewidth,keepaspectratio]{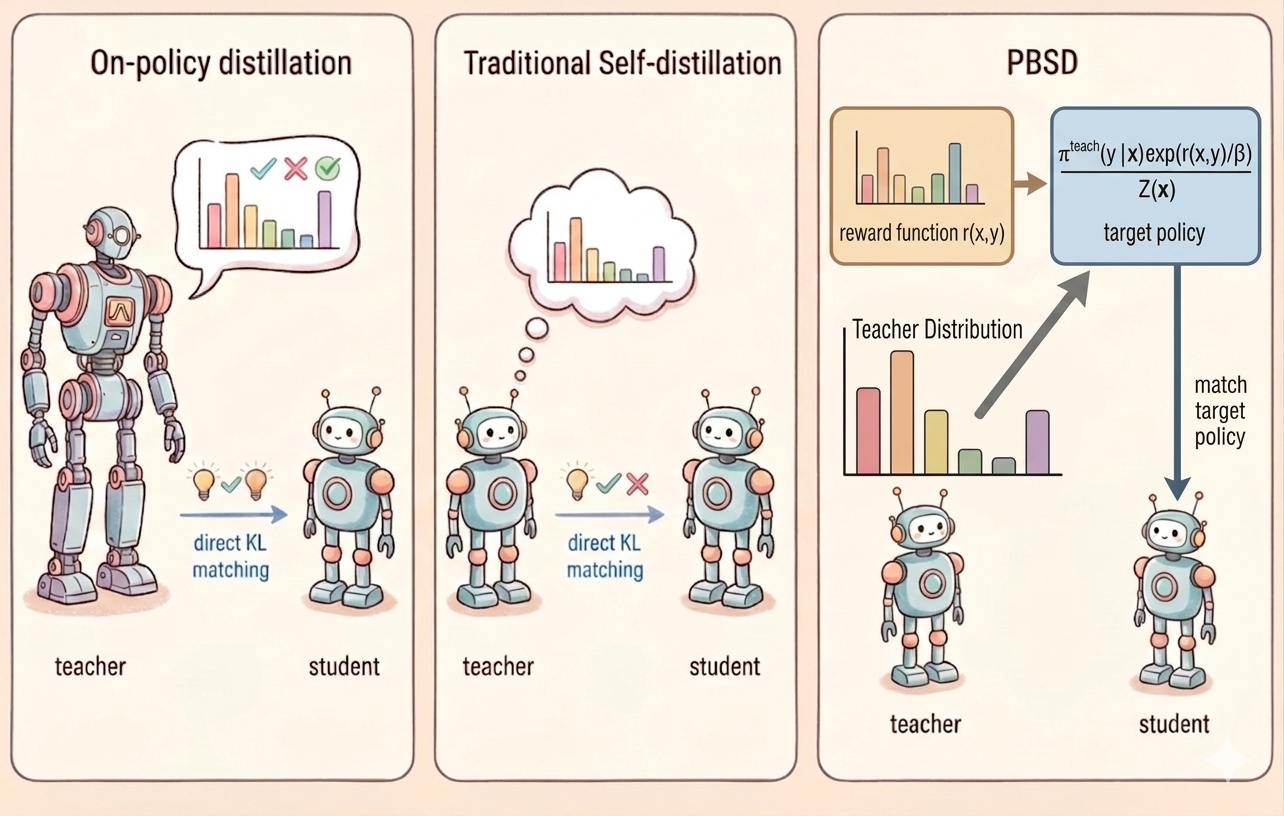}
    \caption{Comparison of three on-policy distillation paradigms. Left: Standard on-policy distillation relies on a stronger external teacher, training the student via direct KL-based distribution matching. Middle: Traditional self-distillation replaces the external teacher with the same base model under a privileged context and retains direct KL matching toward the induced teacher distribution. Right: Our proposed PBSD moves beyond direct KL matching by combining the teacher distribution with a reward function to construct a reward-reweighted target policy for the student to learn, which provides more stable and exploratory supervision for the student.}
    \label{fig_tsd_vs_DPOSD}
\end{figure}

Despite its promise, this KL-based formulation of on-policy self-distillation suffers from two key limitations. First, directly optimizing a KL divergence toward the context-augmented model is unstable and can actively degrade reasoning performance over the course of training, as observed in recent analyses of both self-distillation and on-policy distillation~\citep{kim2026does,fu2026revisiting,li2026rethinkingopd}. Specifically, KL matching tends to suppress epistemic verbalization (namely, the model's explicit expression of uncertainty, hesitation, self-checking, and error correction), yielding reasoning traces that are shorter but undesirably overconfident. Second, as illustrated in Figure~\ref{fig_tsd_vs_DPOSD}, treating a context-augmented model as a teacher is a fundamentally strong assumption. Unlike standard OPD, this teacher shares the exact parameters of the student and differs only in its prompt. Thus, the resulting supervision often lacks the diversity and exploratory value provided by a genuinely stronger external model~\citep{li2026rethinkingopd}. Together, these observations motivate us to move beyond fixed-teacher KL matching. Instead, we seek a more robust target distribution that better preserves exploratory reasoning and admits more stable optimization.

\begin{table}[t]
\centering
\small
\setlength{\tabcolsep}{4pt}
\renewcommand{\arraystretch}{1.10}
\caption{Comparison of post-training paradigms. ``Dense Supervision'' denotes the use of response- or token-level learning signals as opposed to sparse outcome-level feedback, and  ``No External Teacher/Scorer'' indicates that training operates entirely without an auxiliary teacher model or reward/scoring model. Under these criteria, PBSD is the only method that achieves on-policy optimization, dense supervision, token efficiency, and reward-aware optimization without relying on an external teacher or scorer.}
\label{tab:method-comparison}
\resizebox{0.92\linewidth}{!}{
\begin{tabular}{lccccc}
\toprule
\textbf{Method} & \textbf{On-Policy} & \textbf{Dense Supervision} & \textbf{No External Teacher/Scorer} & \textbf{Token Efficiency} & \textbf{Reward-Aware} \\
\midrule
SFT & \xmark & \cmark & \cmark & \cmark & \xmark \\
GRPO & \cmark & \xmark & \xmark & \xmark & \cmark \\
OPD & \cmark & \cmark & \xmark & \cmark & \xmark \\
OPSD & \cmark & \cmark & \cmark & \cmark & \xmark \\
\textbf{PBSD (Ours)} & \cmark & \cmark & \cmark & \cmark & \cmark \\
\bottomrule
\end{tabular}
}
\end{table}

In this work, we propose \textbf{P}reference-\textbf{B}ased \textbf{S}elf-\textbf{D}istillation (\textbf{PBSD}) by revisiting on-policy self-distillation through a reward-regularized lens. Rather than relying solely on KL divergence, we introduce a teacher-anchored objective (i.e., Eq.~(\ref{eq:method-reward-regularized-objective})) that augments KL matching with reward maximization. Under this formulation, the student is encouraged to stay close to the teacher while shifting probability mass toward responses with higher latent reward. Crucially, this objective admits an analytic optimum in which the target policy is a reward-reweighted version of the teacher distribution, rather than the teacher distribution itself. Motivated by this optimal target, PBSD optimizes the preference gaps between teacher and student samples while maintaining on-policy student sampling. We further provide a local statistical analysis of the induced preference-learning problem, clarifying when contextual self-distillation can yield more favorable comparison geometry than external-teacher distillation in our setting. As summarized in Table~\ref{tab:method-comparison}, PBSD helps bridge the gap between classic post-training methods by retaining the rich teacher-derived signal of self-distillation while aligning the optimization process with reward-aware policy improvement.

We summarize our main contributions as follows:
\begin{itemize}
    \item We introduce a reward-regularized objective for self-distillation that balances closeness to the teacher with reward maximization. We show that its closed-form analytic optimum, obtained by reweighting the teacher distribution via latent rewards, improves over the original teacher distribution under this objective.
    \item We propose \textbf{PBSD}, a novel preference-based on-policy self-distillation framework. Furthermore, we provide a local statistical analysis of the induced preference-learning problem, clarifying conditions under which contextual self-distillation yields more favorable comparison geometry than external-teacher distillation.
    \item We empirically evaluate our proposed PBSD on mathematical reasoning and tool-use benchmarks across multiple model scales. In our comparisons, PBSD achieves the strongest overall performance, showing improved training stability over prior self-distillation baselines while retaining the token-efficiency benefits of self-distillation.
\end{itemize}

\section{Related Work}
\label{sec:related-work}

\textbf{On-policy distillation.} On-policy distillation (OPD) trains the student on its own sampled trajectories while querying a stronger teacher on those visited states, thereby preserving the on-policy nature of RL-style optimization while replacing sparse rewards with dense token-level supervision~\citep{agarwal2024opd,song2026survey,fu2026revisiting}. Recent work extends this paradigm to reasoning, RLVR, and domain-specific agent settings~\citep{zhao2026self,yang2026self,li2026sample,afsharrad2026policy}. However, standard OPD still depends on an external teacher and often suffers from teacher--student mismatch in vocabulary, output space, and distributional support~\citep{li2026rethinkingopd,song2026survey}.

\textbf{Self-distillation.} Recent self-distillation methods remove the external teacher by reusing the same base model as both teacher and student, typically by constructing the teacher from additional context, demonstrations, or stronger decoding-time signals~\citep{zhao2026self,shenfeld2026self,sang2026policy,hubotter2026reinforcement,zhang2026embarrassingly,chen2026onesearch}. Despite their differences, these methods still largely optimize KL-based teacher matching. PBSD differs from this line by starting from a reward-regularized objective whose target policy is a reward-reweighted teacher distribution rather than the teacher distribution itself.

\textbf{Preference optimization.} A parallel line of work studies reward-regularized preference optimization through RLHF and DPO-style objectives~\citep{zhu2023principled,rafailov2023direct,zhu2024iterative,pang2024iterative,qi2024online,shi2025crucial,tu2025enhancing}. PBSD borrows the logistic preference structure of this literature, but the setting and data construction are different: instead of using externally annotated preference pairs or general online preference updates, PBSD constructs teacher--student preference pairs specifically for contextual self-distillation and uses them to optimize a self-distillation objective induced by the reward-regularized formulation above. Appendix~\ref{app:related-work} provides a broader discussion.

\section{Methodology}
\label{sec:method}
In this section, we first derive a reward-aware objective for on-policy distillation by augmenting KL-based matching with reward maximization, and then show how to optimize the resulting objective through preference-based learning.
\subsection{A Reward-Regularized Objective for On-Policy Distillation}
\label{subsec:reward-regularized-motivation}

Formally, let \(x\) denote an input prompt and let the student model be parameterized by \(\pi_{\theta}(y \mid x)\), where \(y = (y_1, \dots, y_T)\) is an output sequence of length \(T\). Under the standard autoregressive factorization,
\begin{equation}
\pi_{\theta}(y \mid x)
= \prod_{t=1}^{T} \pi_{\theta}(y_t \mid x, y_{<t}).
\end{equation}

Let the teacher policy be denoted by \(\pi^{\mathrm{teach}}(\cdot\mid x)\), omitting parameters because the teacher is fixed during optimization. Classical on-policy distillation often trains the student to match the teacher distribution through KL divergence. A common formulation minimizes the reverse KL from the student to the teacher:
\begin{equation}
\min_{\pi_{\theta}}
\;
\mathbb{E}_{x\sim\mathcal{D}}
\left[
D_{\mathrm{KL}}\!\left(
\pi_{\theta}(\cdot\mid x)
\,\|\, 
\pi^{\mathrm{teach}}(\cdot\mid x)
\right)
\right],
\label{eq:method-soft-selection-kl}
\end{equation}
which has been used in recent on-policy distillation methods \citep{agarwal2024opd,lu2025onpolicydistillation,zhao2026self,hubotter2026reinforcement}. Under this objective, the teacher distribution itself is treated as the target to be learned. The limitation is that pure KL matching does not distinguish which teacher-supported responses are more useful for the downstream objective. To address this issue, we instead optimize KL matching together with reward maximization. Let \(r(x,y)\) denote the latent target reward. Aggregated over inputs \(x\sim\mathcal{D}\), we consider
\begin{equation}
\max_{\pi_{\theta}(\cdot\mid x)}
\;
\mathbb{E}_{x\sim\mathcal{D}} \mathbb{E}_{y\sim\pi_{\theta}(\cdot\mid x)}[r(x,y)]
-
\beta D_{\mathrm{KL}}\!\left(
\pi_{\theta}(\cdot\mid x)\,\|\,\pi^{\mathrm{teach}}(\cdot\mid x)
\right),
\label{eq:method-reward-regularized-objective}
\end{equation}
where \(\beta>0\) controls the strength of the KL regularization. The KL term keeps the policy close to \(\pi^{\mathrm{teach}}\), while the reward term favors responses with larger latent reward. This differs from classical RLHF in an important way: in standard RLHF, the KL regularizer is typically defined with respect to the initial base model as the reference policy, whereas in our setting the reference policy is the teacher policy \(\pi^{\mathrm{teach}}\), which may be either an external teacher or an internal teacher constructed from the same base model. When \(\beta\) is large, the objective behaves like a soft reward-weighted selection over teacher-supported responses. In this sense, pure KL matching is a special case that ignores the reward-dependent importance of each response, whereas Eq.~(\ref{eq:method-reward-regularized-objective}) redistributes probability mass within the teacher-supported region according to reward. This distinction is central to our view of on-policy distillation: the teacher should define useful support and inductive bias, but it should not be treated as a final target to be copied uniformly.

\begin{proposition}[Optimal reward-tilted policy]
\label{prop:method-reward-tilted-policy}
For each fixed input \(x\), the optimizer of Eq.~(\ref{eq:method-reward-regularized-objective}) is the reward-tilted teacher distribution
\begin{equation}
\pi^\star(y\mid x)
=
\frac{
\pi^{\mathrm{teach}}(y\mid x)\exp(r(x,y)/\beta)
}{
Z(x)
},
\qquad
Z(x)=
\sum_y
\pi^{\mathrm{teach}}(y\mid x)\exp(r(x,y)/\beta).
\label{eq:method-reward-tilted-policy}
\end{equation}
Equivalently, the latent reward can be written in terms of the optimal policy and the teacher up to an \(x\)-dependent additive constant:
\begin{equation}
r(x,y)
=
\beta
\log\frac{\pi^\star(y\mid x)}{\pi^{\mathrm{teach}}(y\mid x)}
+
\beta\log Z(x).
\label{eq:method-likelihood-ratio}
\end{equation}
\end{proposition}
The derivation is deferred to Appendix~\ref{app:method-reward-regularized-details}. Thus, the desired policy does not merely copy the teacher. Instead, it takes the teacher distribution as a reference measure and reweights it by the reward-dependent factor \(\exp(r(x,y)/\beta)\). The following proposition formalizes that this reward-tilted solution is never worse than directly using the teacher under the same reward-regularized objective.
\begin{proposition}[Reward-tilted policy improves over the teacher]
\label{prop:method-reward-tilted-improvement}
Let \(F\) denote the population version of Eq.~(\ref{eq:method-reward-regularized-objective}) aggregated over \(x\sim\mathcal{D}\). Then the optimizer \(\pi^\star\) in Eq.~(\ref{eq:method-reward-tilted-policy}) satisfies
\begin{equation}
F(\pi^\star)\ge F(\pi^{\mathrm{teach}}),
\end{equation}
with strict inequality whenever \(r(x,y)\) is non-constant over teacher-supported responses on a set of inputs with positive measure.
\end{proposition}
The proof is deferred to Appendix~\ref{app:method-reward-regularized-details}. This observation is important for on-policy distillation: the teacher is often not an oracle and therefore should not be treated as the final target. Instead, the teacher should serve as a support distribution, while latent reward information reweights probability mass toward more valuable behavior. This perspective matches the motivation in Section~\ref{sec:introduction}: on-policy distillation should not uniformly copy the teacher, but should selectively amplify higher-value teacher-induced responses. The remaining question is how to optimize this reward-aware target in practice when the reward itself is unobserved.

\subsection{Solving the Objective via Preference-Based Optimization}
\label{subsec:dpo-induced}

The previous subsection defines the desired target policy, but that policy depends on a latent reward that is not directly observed. To make the objective optimizable in practice, we instantiate it through preference-based learning. Specifically, we compare a response sampled from the better-conditioned teacher with an on-policy response sampled from the current student, and use the induced preference relation to optimize the student toward the reward-reweighted teacher policy.

Our teacher is a \emph{better-conditioned} policy rather than a larger external model. For each input \(x\), we assume an additional context \(\boldsymbol{c}\) that is available only when constructing the teacher signal. In prior self-distillation work, this context is typically instantiated as privileged per-instance information, such as expert demonstrations or reference reasoning traces in reasoning tasks~\citep{zhao2026self,sang2026policy}, tool or API information in structured action-generation settings~\citep{shenfeld2026self}, retrieved evidence or search traces in retrieval- or search-augmented settings~\citep{chen2026onesearch}, or more general teacher-only feedback signals~\citep{ding2026hdpo,hubotter2026reinforcement}. We denote the resulting teacher by
\begin{equation}
\pi^{\mathrm{teach}}(y\mid x)
:=
\pi(y\mid x,\boldsymbol{c}),
\end{equation}
while the student remains \(\pi_{\theta}(y\mid x)\). The teacher and student may share the same base model; their distinction comes from the conditioning information, not necessarily from model capacity.

Using the optimal-policy identity induced by the reward-regularized KL objective, the latent reward can be represented by the log-ratio between the optimized policy and a reference policy, up to an input-dependent constant. In our setting, we use the better-conditioned teacher as the reference policy and optimize the student. For a preferred response \(y^{+}\) and a less preferred response \(y^{-}\), the resulting preference margin is
\begin{equation}
m_{\theta}(x,y^{+},y^{-})
=
\beta
\left[
\log
\frac{\pi_{\theta}(y^{+}\mid x)}
{\pi^{\mathrm{teach}}(y^{+}\mid x)}
-
\log
\frac{\pi_{\theta}(y^{-}\mid x)}
{\pi^{\mathrm{teach}}(y^{-}\mid x)}
\right].
\label{eq:method-dposd-margin}
\end{equation}
This margin induces, under a Bradley-Terry preference model, the probability that the teacher-generated response is preferred over the student-generated response:
\begin{equation}
P_{\theta}(y^{+}\succ y^{-}\mid x)
=
\sigma\!\left(
m_{\theta}(x,y^{+},y^{-})
\right).
\label{eq:method-dposd-pref-prob}
\end{equation}
Here \(\sigma(z):=1/(1+\exp(-z))\) denotes the logistic sigmoid function.
Operationally, each PBSD update uses teacher--student comparisons collected from the current student rollout and then optimizes the pairwise likelihood on that sampled batch. Writing \(\pi^{\mathrm{old}}\) for the behavior policy used to generate the current student responses, we maximize the corresponding sampled pairwise log-likelihood,
\begin{equation}
\max_{\theta}
\;
\mathbb{E}_{(x,\boldsymbol{c})\sim\mathcal{D}}
\mathbb{E}_{
y^{+}\sim\pi^{\mathrm{teach}}(\cdot\mid x),
y^{-}\sim\pi^{\mathrm{old}}(\cdot\mid x)}
\left[
\log\sigma\!\left(
m_{\theta}(x,y^{+},y^{-})
\right)
\right].
\label{eq:method-dposd-objective}
\end{equation}
Equivalently, the implementation minimizes the negative log-likelihood in Eq.~(\ref{eq:method-dposd-objective}). Here \(y^{+}\) is generated by the better-conditioned teacher and \(y^{-}\) is generated on policy by the current student behavior policy \(\pi^{\mathrm{old}}\). During the gradient step, the sampled pair is treated as fixed while gradients flow only through the optimized student policy \(\pi_{\theta}\). Maximizing Eq.~(\ref{eq:method-dposd-objective}) increases the relative probability that the student assigns to teacher-generated responses over its own current responses. This online construction makes the preference signal adaptive: the teacher rollout provides an attraction term toward better-supported behavior, while the student rollout exposes current failure modes that should be downweighted. PBSD therefore preserves the logistic structure of DPO, but uses it as a self-distillation mechanism induced by the reward-reweighted objective in Section~\ref{subsec:reward-regularized-motivation}.
Unlike standard DPO, the reference policy in PBSD is the contextual teacher rather than the initial base model, and the positive/negative responses are generated from the teacher and current student respectively. This distinction is central to our use of preference optimization as an online self-distillation mechanism rather than a generic post-hoc alignment objective.

The token-level effect of this objective is explicit from its gradient. For a sampled triple \((x_i,y_i^{+},y_i^{-})\), define
\begin{equation}
\ell_i(\theta)
=
-
\log\sigma\!\left(m_{\theta}(x_i,y_i^{+},y_i^{-})\right).
\label{eq:method-sample-loss}
\end{equation}
Since the teacher policy is fixed, differentiating only through the student policy gives
\begin{align}
\nabla_{\theta}\ell_i(\theta)
&=
-
\beta
\sigma\!\left(-m_{\theta}(x_i,y_i^{+},y_i^{-})\right)
\left[
\nabla_{\theta}\log\pi_{\theta}(y_i^{+}\mid x_i)
-
\nabla_{\theta}\log\pi_{\theta}(y_i^{-}\mid x_i)
\right]
\nonumber\\
&=
-
\beta
\sigma\!\left(-m_{\theta}(x_i,y_i^{+},y_i^{-})\right)
\left[
\sum_{k=1}^{|y_i^{+}|}
\nabla_{\theta}
\log\pi_{\theta}(y_{i,k}^{+}\mid x_i,y_{i,<k}^{+})
-
\sum_{k=1}^{|y_i^{-}|}
\nabla_{\theta}
\log\pi_{\theta}(y_{i,k}^{-}\mid x_i,y_{i,<k}^{-})
\right].
\label{eq:method-token-gradient}
\end{align}
Thus, gradient descent on \(\ell_i\) increases the likelihood of teacher-generated tokens while decreasing the likelihood of the student-generated negative sample, with an adaptive weight \(\sigma(-m_{\theta}(x_i,y_i^{+},y_i^{-}))\) that becomes larger when the current student assigns insufficient relative preference to \(y_i^{+}\). The full online training procedure is summarized in Algorithm~\ref{alg:dposd}.

\begin{figure}[h]
\centering
\refstepcounter{algorithm}\label{alg:dposd}
\fbox{
\begin{minipage}{0.95\linewidth}
\small
\textbf{Algorithm \thealgorithm: PBSD: Preference-Based Online Self-Distillation}

\textbf{Input:} Training set $\mathcal{D}$ of pairs $(x, \boldsymbol{c})$; student policy $\pi_{\theta}(y \mid x)$; teacher policy $\pi^{\mathrm{teach}}(y\mid x):=\pi(y \mid x, \boldsymbol{c})$; temperature $\beta$; learning rate $\eta$; total training steps $T$.

\textbf{Output:} Updated student parameters $\theta$.

\textbf{For} $t = 1, \dots, T$:

\hspace*{1em} Sample a mini-batch $\mathcal{B} = \{(x_i, \boldsymbol{c}_i)\}_{i=1}^{B}$ from $\mathcal{D}$.

\hspace*{1em} \textbf{For each} $(x_i, \boldsymbol{c}_i) \in \mathcal{B}$:

\hspace*{2em} Generate a student response $y_i^{-} \sim \pi_{\theta}(\cdot \mid x_i)$.

\hspace*{2em} Generate a teacher response $y_i^{+} \sim \pi^{\mathrm{teach}}(\cdot \mid x_i)$.

\hspace*{2em} Compute the PBSD margin $m_{\theta}(x_i, y_i^{+}, y_i^{-})$ using Eq.~(\ref{eq:method-dposd-margin}).

\hspace*{2em} Compute the pairwise loss $\mathcal{L}_i = -\log \sigma\!\left(m_{\theta}(x_i, y_i^{+}, y_i^{-})\right)$.

\hspace*{1em} Average over the mini-batch:
\[
\mathcal{L}_{\mathrm{PBSD}}=\frac{1}{B}\sum_{i=1}^{B} \mathcal{L}_i.
\]

\hspace*{1em} Update the student policy:
\[
\theta \leftarrow \theta - \eta \nabla_{\theta}\mathcal{L}_{\mathrm{PBSD}}.
\]
\end{minipage}
}
\caption{PBSD training procedure.}
\end{figure}

\section{Theoretical Analysis of PBSD}
\label{sec:theory}

We analyze online PBSD from a statistical perspective. The positive response is drawn from a fixed context-augmented teacher, while the negative response is drawn on-policy from the current student. Our goal is to understand what properties of teacher-generated positives lead to stronger statistical conditions for preference alignment. This perspective is closely related to recent theoretical studies of RLHF and pairwise preference learning \citep{zhu2024iterative,zhu2023principled}, but here we specialize the analysis to the online PBSD objective induced by contextual self-distillation. The details of optimization induction can be found in Appendix~\ref{app:theory-objective-details}; here we focus on the sample-complexity bound of our problem and its interpretation.

\subsection{Pairwise MLE and Informative Comparisons}
\label{subsec:theory-mle}

We now view the PBSD empirical objective as a pairwise MLE, where the local information in a logistic comparison objective is determined by the Hessian of the empirical negative log-likelihood. For the \(i\)-th sampled pair \((x_i,y_i^+,y_i^-)\), let \(\ell_i(\theta)\) denote the sample loss defined in Eq.~(\ref{eq:method-sample-loss}). The empirical negative log-likelihood is
\[
\widehat{\mathcal{L}}_n(\theta)
:=
\frac{1}{n}\sum_{i=1}^n \ell_i(\theta).
\]
The empirical MLE is
\begin{equation}
\widehat{\theta}_n
\in
\arg\min_{\theta}\widehat{\mathcal{L}}_n(\theta).
\label{eq:empirical-mle}
\end{equation}

To expose the local statistical structure, we only introduce sample-indexed notation for the score-gap direction,
\[
d_i(\theta)
:=
\nabla_\theta \log \pi_\theta(y_i^+\mid x_i)
-
\nabla_\theta \log \pi_\theta(y_i^-\mid x_i).
\]
Then
\[
\nabla_\theta m_\theta(x_i,y_i^+,y_i^-)
=
\beta d_i(\theta).
\]
The sample loss \(\ell_i(\theta)\) is generally nonconvex for a neural policy. We therefore focus on its Gauss--Newton component, obtained by differentiating the logistic loss with respect to the margin and keeping the resulting outer-product term. This yields the local information contribution
\begin{equation}
G_i(\theta)
=
\beta^2
\sigma\!\left(m_\theta(x_i,y_i^+,y_i^-)\right)
\left(1-\sigma\!\left(m_\theta(x_i,y_i^+,y_i^-)\right)\right)
d_i(\theta)d_i(\theta)^\top .
\label{eq:sample-local-hessian}
\end{equation}
Averaging over the \(n\) pairs yields the empirical Gauss--Newton information matrix,
\begin{equation}
\widehat{H}_n(\theta)
=
\frac{1}{n}\sum_{i=1}^n G_i(\theta).
\label{eq:dposd-local-hessian}
\end{equation}
For the original nonlinear policy, Eq.~(\ref{eq:dposd-local-hessian}) is the Gauss--Newton component of the Hessian. It is the local statistical object that controls the sample complexity of pairwise estimation. The detailed derivations for this subsection are deferred to Appendix~\ref{app:theory-objective-details}.

\begin{theorem}[Local MLE complexity for PBSD]
\label{prop:dposd-local-mle}
Suppose that, within a local neighborhood of \(\theta^\star\), the pairwise logistic model induced by the PBSD loss is well specified and the Gauss--Newton Hessian in Eq.~(\ref{eq:dposd-local-hessian}) is locally stable, and define
\begin{equation}
\theta^\star
\in
\arg\min_{\theta}\,
\mathcal{L}(\theta),
\qquad
\mathcal{L}(\theta)
:=
\mathbb{E}[\ell_i(\theta)].
\label{eq:theta-star-definition}
\end{equation}
Assume also that the score-gap features are bounded and that \(\lambda_{\min}(\widehat{H}_n(\theta^\star))>0\). Then, with probability at least \(1-\delta\), the local MLE \(\widehat{\theta}_n\) satisfies
\begin{equation}
\left\|
\widehat{\theta}_n-\theta^\star
\right\|_2
\le
C\sqrt{
\frac{d+\log(1/\delta)}
{n\,\lambda_{\min}(\widehat{H}_n(\theta^\star))}
}.
\label{eq:dposd-mle-euclidean}
\end{equation}
Here \(d\) is the local parameter dimension and \(C>0\) is an absolute constant depending only on the boundedness constants.
\end{theorem}
The proof is deferred to Appendix~\ref{app:theory-mle-details}, where we also justify why the required local conditions are mild in our setting. We emphasize that Theorem~\ref{prop:dposd-local-mle} is a finite-sample, data-dependent local characterization: it shows how the local sample complexity is governed by the smallest eigenvalue of the empirical information matrix. For the Hessian estimate, each comparison pair contributes
\begin{equation}
\beta^2
\sigma\!\left(m_{\theta^\star}(x_i,y_i^+,y_i^-)\right)
\left(1-\sigma\!\left(m_{\theta^\star}(x_i,y_i^+,y_i^-)\right)\right)
d_i(\theta^\star)d_i(\theta^\star)^\top .
\label{eq:single-pair-information}
\end{equation}
Beyond this finite-sample characterization, Appendix~\ref{app:theory-mle-details} also gives a strengthened local consistency result. Under a frozen local comparison model, bounded score gaps, and a positive full population-information condition \(\lambda_{\min}(H^\star)\ge \mu>0\), the empirical information is bounded away from zero with high probability for all sufficiently large \(n\). Substituting this estimate into the finite-sample bound yields an explicit vanishing local rate of order \(O(n^{-1/2})\).
\paragraph{Context-Augmented Teacher vs. External Teacher.}
Thus a useful pair must satisfy two complementary conditions. First, from the perspective of self-distillation, the positive samples should come from a distribution that is more diverse than the current student distribution, so that the induced score-gap directions \(d_i(\theta^\star)\) span informative directions beyond those already covered by the student. If the teacher responses collapse onto a narrow region already represented by the student, then the outer-product terms \(d_i(\theta^\star)d_i(\theta^\star)^\top\) contribute little new geometric information. Second, the logistic curvature weight \(\sigma(m_{\theta^\star})(1-\sigma(m_{\theta^\star}))\) requires the teacher--student gap to remain moderate. It is maximized when \(m_{\theta^\star}(x_i,y_i^+,y_i^-)=0\), where it equals \(1/4\), and it decays to zero as \(|m_{\theta^\star}(x_i,y_i^+,y_i^-)|\) grows. In contextual self-distillation, this condition is more plausible because the teacher and student are induced from the same base model and therefore remain relatively close in distribution. By contrast, in more general on-policy distillation with an external teacher model, the distribution shift can be substantially larger, which makes overly large margins and saturation more likely. Through Eq.~(\ref{eq:dposd-mle-euclidean}), these two properties jointly improve statistical complexity: richer score-gap directions together with moderate margins strengthen the Hessian, increase \(\lambda_{\min}(\widehat{H}_n(\theta^\star))\), and tighten the estimation bound.

\section{Experiments}

In this section, we empirically verify the effectiveness of PBSD across two task domains. Following prior work, we conduct experiments on mathematical reasoning. In addition, because improving the agentic quality of open-source models is important, we also evaluate the capability of the fine-tuned model on tool use.

\subsection{Experiment Setup}

\paragraph{Models and Tasks.} We experiment with instruct-tuned Qwen models \citep{yang2025qwen3} at three scales: Qwen3-1.7B, Qwen3-4B, and Qwen3-8B. We study two task domains: \textbf{mathematical reasoning} and \textbf{tool use}. Unless otherwise noted, these two domains use the same training pipeline, model initialization, LoRA configuration, optimizer, teacher--student construction, rollout configuration, and checkpoint-selection protocol; they differ only in their evaluation data and metrics. For mathematical reasoning, we use the math subset of OpenThoughts \citep{guha2025openthoughts}, sampling up to 30K problem--solution pairs with chain-of-thought reasoning, and report results on AIME 2024, AIME 2025, and HMMT 2025. For tool use, we follow the ToolAlpaca-based setup of \citet{shenfeld2026self}, using a 4046-example training split and a 94-example test split. Here, ``agentic quality'' refers to the model's ability to map user intent to correct multi-step or action-oriented behavior, a capability that is central to recent work on LLM agents and open-source tool-using systems \citep{wang2026openclaw,wang2026skillsd}. Detailed dataset statistics are provided in Appendix~\ref{app:exp-datasets}.

\paragraph{Baselines and Metrics.} We compare PBSD with six baselines: SFT, GRPO\citep{shao2024deepseekmath}, DAPO \citep{yu2025dapo}, OPSD \citep{zhao2026self}, SDFT \citep{shenfeld2026self}, and SRPO \citep{li2026sample}. DAPO is a large-scale RLVR recipe that strengthens GRPO-style training through decoupled clipping and dynamic sampling. SDFT is a demonstration-conditioned self-distillation method that enables on-policy learning directly from demonstrations. SRPO is a hybrid on-policy objective that routes correct samples to GRPO-style reward optimization and failed samples to self-distillation-based correction. The training setup is shared across the two task domains, but the evaluation protocol is task-specific. For mathematical reasoning, we report \textbf{\textit{Avg@12}}, namely the average accuracy over 12 sampled responses per question. For tool use, we report top-1 accuracy on the test split. Detailed evaluation protocols for both tasks are provided in Appendix~\ref{app:exp-eval}.

\paragraph{Implementation Details.} Across all methods and both task domains, we fine-tune Qwen3 instruct models with LoRA (rank 64 and \(\alpha=128\)) on 8 H100 GPUs using AdamW, bfloat16 precision, gradient checkpointing, and FlashAttention~2. We fix the teacher policy to the initial checkpoint during PBSD training for stability. Following OPSD, all trainable methods are trained for 500 steps and evaluated every 50 steps. Unless otherwise noted, we report the latest checkpoint results at the end of this shared 500-step budget. Appendix~\ref{app:exp-long-horizon} additionally reports the best checkpoint within 500 steps together with a longer-horizon analysis up to 1{,}000 steps for PBSD. Detailed training configurations, including the shared setup and method-specific hyperparameters for SFT, GRPO, DAPO, OPSD, SDFT, SRPO, and PBSD, are provided in Appendix~\ref{app:exp-training}.
\subsection{Main Results}
\begin{table}[t]
\centering
\footnotesize
\caption{Performance comparison across mathematical reasoning and tool use for Qwen3 models. Under \textbf{Math}, we report Avg@12 using the standard Qwen3 sampling configuration (temperature 1.0, maximum generation length 38k). Under \textbf{Tool Use}, we report top-1 accuracy on the tool-use test set. Numbers in parentheses denote standard deviation over three random seeds. All trainable methods are trained for 500 steps and evaluated every 50 steps. This main table reports the latest checkpoint results at the end of the shared 500-step budget; Appendix~\ref{app:exp-long-horizon} additionally reports the best checkpoint within 500 steps and longer-horizon checkpoints for PBSD. Within each model scale, the best result in each column is shown in \textbf{bold}, and the second-best result is \underline{underlined}.}
\label{tab:main_math_results}
\setlength{\tabcolsep}{4pt}
\renewcommand{\arraystretch}{0.98}
\resizebox{0.76\linewidth}{!}{
\begin{tabular}{llcccc|c}
\toprule
\multirow{2}{*}{\textbf{Model}} & \multirow{2}{*}{\textbf{Method}} & \multicolumn{4}{c|}{\textbf{Math}} & \multicolumn{1}{c}{\textbf{Tool Use}} \\[2pt]
\cline{3-7}
 & & \rule{0pt}{2.4ex}\textbf{AIME24} & \textbf{AIME25} & \textbf{HMMT25} & \textbf{Average} & \textbf{Acc.} \\
\midrule
\textit{Qwen3-8B} & Base (Instruct) & 75.6 (0.5) & 65.3 (0.4) & 43.6 (0.2) & 61.5 & 61.3 (0.5) \\
 & + SFT & 71.9 (1.5) & 63.9 (0.2) & 42.4 (0.3) & 59.4 & 61.7 (1.7) \\
 & + GRPO & 76.1 (0.6) & 69.0 (0.1) & \textbf{46.3} (0.5) & 63.8 & \underline{68.8} (2.7) \\
 & + DAPO & 76.0 (0.9) & 68.7 (1.0) & 46.0 (0.7) & 63.6 & 67.7 (0.5) \\
 & + OPSD & \underline{77.3} (0.3) & \underline{70.1} (0.6) & 45.1 (0.6) & \underline{64.2} & 65.6 (1.8) \\
 & + SDFT & 75.9 (0.7) & 69.1 (0.7) & 44.2 (0.6) & 63.1 & 62.8 (1.7) \\
 & + SRPO & 77.1 (0.3) & 69.6 (1.5) & 43.9 (0.7) & 63.5 & 62.1 (2.5) \\
\rowcolor{gray!8}
\textit{} & + PBSD & \textbf{78.4} (0.3) & \textbf{71.0} (0.1) & \underline{46.1} (0.2) & \textbf{65.2} & \textbf{72.0} (1.3) \\
\midrule
\textit{Qwen3-4B} & Base (Instruct) & 74.5 (0.3) & 66.1 (0.2) & 42.0 (0.3) & 60.9 & 45.7 (0.9) \\
 & + SFT & 71.3 (0.6) & 64.3 (0.8) & 43.4 (0.1) & 59.7 & 51.1 (1.7) \\
 & + GRPO & 75.7 (0.1) & 67.8 (0.6) & 44.4 (0.8) & 62.6 & \underline{58.9} (1.3) \\
 & + DAPO & 76.1 (0.6) & 67.7 (0.3) & 44.3 (0.7) & 62.7 & 48.2 (2.2) \\
 & + OPSD & \underline{76.3} (0.8) & \underline{68.1} (0.3) & \textbf{46.0} (0.6) & \underline{63.5} & 53.9 (1.0) \\
 & + SDFT & 75.6 (0.3) & 67.1 (0.3) & 44.7 (0.2) & 62.5 & 41.8 (1.3) \\
 & + SRPO & \underline{76.3} (0.3) & 67.8 (0.4) & 45.5 (0.3) & 63.2 & 46.1 (2.0) \\
\rowcolor{gray!8}
\textit{} & + PBSD & \textbf{77.3} (0.1) & \textbf{69.0} (0.1) & \underline{45.6} (0.3) & \textbf{64.0} & \textbf{60.6} (1.7) \\
\midrule
\textit{Qwen3-1.7B} & Base (Instruct) & 51.4 (0.2) & 37.2 (0.5) & 23.1 (0.1) & 37.3 & 36.9 (1.0) \\
 & + SFT & 49.9 (0.3) & 36.6 (0.5) & 22.8 (0.2) & 36.4 & 39.7 (0.5) \\
 & + GRPO & 50.6 (0.5) & 38.4 (0.1) & 23.5 (0.3) & 37.5 & 42.9 (0.5) \\
 & + DAPO & 51.5 (0.7) & 38.6 (1.4) & 23.5 (0.3) & 37.9 & \underline{43.6} (0.9) \\
 & + OPSD & 57.1 (0.8) & 43.3 (0.6) & 29.3 (0.3) & 43.2 & 37.2 (1.7) \\
 & + SDFT & 57.0 (0.3) & 43.0 (0.1) & 28.7 (0.3) & 42.9 & 38.3 (1.7) \\
 & + SRPO & \underline{58.1} (0.2) & \underline{44.1} (0.3) & \underline{29.4} (0.5) & \underline{43.9} & 36.9 (1.3) \\
\rowcolor{gray!8}
 & + PBSD & \textbf{58.5} (0.5) & \textbf{44.4} (0.2) & \textbf{30.0} (0.2) & \textbf{44.3} & \textbf{44.7} (1.7) \\
\bottomrule
\end{tabular}
}
\end{table}

Table~\ref{tab:main_math_results} reports the main results across both mathematical reasoning and tool use for three Qwen3 model scales. PBSD consistently improves over the base instruct model and achieves the strongest average result at all three model scales. On Qwen3-8B, PBSD reaches a math average of \(65.2\), improving over the base instruct model by \(3.7\) points and over the strongest baseline OPSD by \(1.0\) point. On Qwen3-4B, PBSD attains a math average of \(64.0\), again outperforming both the base model (\(60.9\)) and OPSD (\(63.5\)). At the 1.7B scale, PBSD now delivers the best result on all three math benchmarks and reaches the strongest overall average of \(44.3\), ahead of SRPO (\(43.9\)) and OPSD (\(43.2\)). On tool use, PBSD also achieves the strongest result at every model scale, reaching \(72.0\), \(60.6\), and \(44.7\) for Qwen3-8B, Qwen3-4B, and Qwen3-1.7B, respectively. These results show that pairwise self-distillation with a contextual teacher provides a more effective optimization signal than directly matching the teacher distribution or relying on pure reward optimization.

Compared with prior baselines, the gains from PBSD are clearest in overall consistency across tasks and scales. On Qwen3-8B, PBSD achieves the best result on AIME24, AIME25, the overall math average, and tool use, while remaining second-best on HMMT25 behind GRPO. On Qwen3-4B, PBSD again gives the strongest average result, with consistent improvements on AIME24, AIME25, and tool use, although OPSD remains slightly better on HMMT25. At the 1.7B scale, PBSD surpasses both OPSD and SRPO across all three math benchmarks and also gives the best tool-use result. Taken together, these trends suggest that PBSD is particularly effective when the student can benefit from teacher-induced positive samples without sacrificing exploration, while still preserving the token-efficiency advantages of self-distillation.
Additional robustness and generalization analyses are reported in Appendix~\ref{app:experiment-details}, including broader-domain transfer on coding and instruction-following, OOD capability diagnostics, sensitivity to privileged context and \(\beta\), larger-teacher comparisons, and additional last-checkpoint and longer-horizon checkpoint analysis.
\subsection{Sample Efficiency and Training Dynamics}

\begin{figure}[t]
\centering
\begin{minipage}[t]{0.45\linewidth}
\centering
\includegraphics[width=\linewidth]{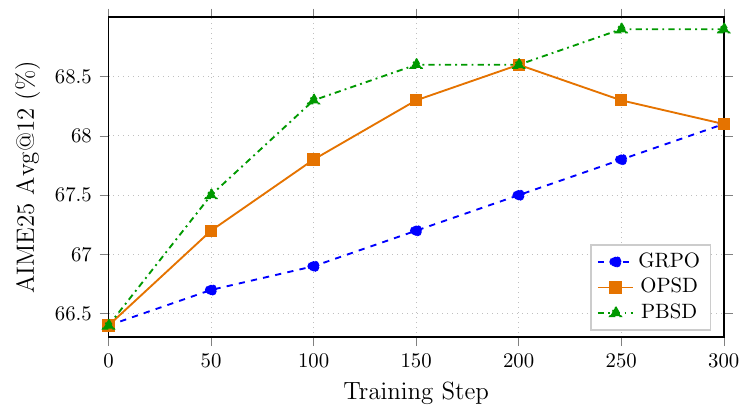}
\par\vspace{1pt}
{\small (A)}
\end{minipage}
\hfill
\begin{minipage}[t]{0.45\linewidth}
\centering
\includegraphics[width=\linewidth]{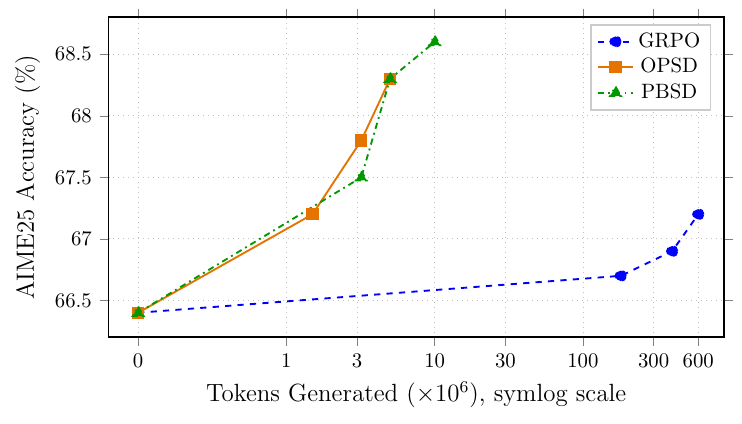}
\par\vspace{1pt}
{\small (B)}
\end{minipage}
\caption{AIME25 training-dynamics comparison among GRPO, OPSD, and PBSD. (A) Avg@12 versus training steps. PBSD remains stable throughout training and achieves the strongest final performance, whereas OPSD peaks early and then degrades. (B) AIME25 accuracy versus cumulative generated tokens. PBSD achieves a favorable accuracy--token trade-off, remaining much more token-efficient than GRPO while staying competitive with OPSD. The x-axis in (B) uses a symlog scale: it is linear from 0 to 1M tokens and logarithmic beyond 1M tokens.}
\label{fig:aime25-ablation-results}
\end{figure}

\paragraph{Stable during training.} Figure~\ref{fig:aime25-ablation-results}A shows that PBSD does not degenerate as training proceeds. Unlike OPSD, which peaks early and then declines, PBSD continues to improve and achieves the strongest final AIME25 performance.

\paragraph{Token efficiency.} Figure~\ref{fig:aime25-ablation-results}B shows that PBSD is token-efficient throughout training. It reaches high final performance without requiring the very large token budget used by RL-based optimization.

We provide additional ablations in Appendix~\ref{app:experiment-details} to further compare prompt-level gains from expert demonstrations and training-time gains from OPSD and PBSD, summarized as follows:

\paragraph{Expert demonstrations correct base-model errors.} Table~\ref{tab:aime25-case-study-template} shows a case-by-case study on 30 selected AIME25 problems. While directly injecting expert demonstrations yields near-perfect problem-level performance, both OPSD and PBSD also recover a substantial portion of the base model's majority-vote errors, indicating that self-distillation can absorb useful supervision from the privileged teacher signal.

\paragraph{PBSD preserves exploration.} Table~\ref{tab:aime24-length-by-problem} compares student and teacher accuracy together with completion lengths from the base student, the demonstration-conditioned teacher, OPSD, and PBSD. In particular, it lets us check whether PBSD preserves a more moderate completion length than the teacher and OPSD while still improving final accuracy.

\paragraph{A fixed teacher is sufficient.} Table~\ref{tab:teacher-update-frequency} shows that keeping the teacher fixed at initialization is already effective. Updating the teacher every 5 steps does not provide a clear benefit, suggesting that a stable teacher signal is more important than a rapidly refreshed reference in our setting.
\vspace{-4pt}
\section{Limitations}
\label{sec:limitations}

PBSD still depends on the quality of the contextual teacher signal. When the privileged context is weak, noisy, or only marginally informative, the induced teacher can also be weak, and the benefit of self-distillation is correspondingly reduced. More broadly, our current experiments focus on reasoning-centric domains and structured tool-use settings, so broader open-ended behaviors still deserve further study. At the same time, this limitation is closely tied to the problem regime we target: contextual self-distillation is precisely challenging because the teacher is not a strong external oracle, but only a better-conditioned version of the same base model. PBSD is designed to improve this regime by reweighting the teacher signal rather than copying it uniformly.

\section{Conclusion}

In this work, we revisited on-policy self-distillation through a reward-regularized lens, establishing that direct KL matching to a context-augmented teacher fundamentally limits both stability and reward awareness. To overcome these bottlenecks, we proposed PBSD, a preference-based self-distillation framework that learns a reward-reweighted target policy rather than uniformly imitating the teacher. We also provided a local statistical analysis clarifying when contextual self-distillation can yield more favorable comparison geometry than external-teacher distillation. Empirically, across multiple model scales on mathematical reasoning and tool-use benchmarks, PBSD consistently achieved the strongest average performance. It successfully combines the token efficiency of standard self-distillation with significantly improved training stability. These results highlight reward-aware self-distillation as a promising paradigm for stable, scalable post-training of reasoning-oriented LLMs.

% Prefer the pre-generated bibliography for stable arXiv/Overleaf compilation.
\IfFileExists{neurips_2026.bbl}{

}{
\bibliographystyle{plainnat}
\bibliography{ref}
}

\clearpage

%%%%%%%%%%%%%%%%%%%%%%%%%%%%%%%%%%%%%%%%%%%%%%%%%%%%%%%%%%%%

\appendix

\clearpage
\section{Additional Discussion of Limitations}
\label{app:limitations}

This appendix briefly expands the discussion in Section~\ref{sec:limitations}. The main limitation of our setting is that PBSD still depends on the quality of the contextual teacher signal. When the additional privileged context is weak or only marginally informative, the induced teacher can also be weak, and the benefit of self-distillation is correspondingly reduced. At the same time, this is precisely the regime that PBSD is designed to improve: the teacher is not a strong oracle but only a better-conditioned version of the same base model, so the goal is not to copy it uniformly, but to extract the more valuable part of its signal through reward-aware reweighting.

\clearpage
\section{Appendix Overview}
This appendix is organized to mirror the logic of the main paper from background, to derivation, to theory, and finally to experimental detail. Appendix~\ref{app:related-work} expands the literature discussion that is abbreviated in the main text. It is divided into three thematic parts: on-policy distillation, self-distillation, and RLHF/DPO-style preference optimization. The goal of this section is to situate PBSD relative to dense teacher-based post-training, contextual self-distillation, and reward-aware preference optimization.

Appendix~\ref{app:method-proofs} provides the derivations that support the methodology section. Section~\ref{app:method-reward-regularized-details} contains two components: first, the derivation of the optimal reward-tilted teacher distribution used in Proposition~\ref{prop:method-reward-tilted-policy}; second, the proof that this reward-tilted solution is no worse than directly using the teacher under the same reward-regularized objective, corresponding to Proposition~\ref{prop:method-reward-tilted-improvement}. This appendix therefore justifies the objective-level motivation for replacing pure KL matching with reward-aware reweighting.

Appendix~\ref{app:theory-proofs} develops the technical details behind the statistical analysis in the main text. Section~\ref{app:theory-objective-details} derives the sample-level gradient and the local Hessian form of the online PBSD objective, which are the ingredients needed to interpret PBSD as a pairwise logistic-style estimation problem. Section~\ref{app:theory-mle-details} then states the local assumptions used in the analysis, explains why these assumptions are mild in our setting, discusses what happens when they are violated, and finally proves the local MLE-style error bound for the induced learning problem.

Appendix~\ref{app:experiment-details} collects the detailed experimental material that supports Section~4 of the main paper. Section~\ref{app:exp-datasets} describes the two task domains, namely mathematical reasoning and tool use, together with the corresponding training and evaluation splits. The data-visualization subsection then gives concrete examples of the processed mathematical reasoning and tool-use instances so that the structure of the training data is explicit. Section~\ref{app:exp-training} documents the shared training setup and the method-specific configurations for SFT, GRPO, DAPO, OPSD, SDFT, SRPO, and PBSD. Section~\ref{app:exp-eval} specifies the decoding and evaluation protocols for both mathematical reasoning and tool use. The case-by-case study subsection explains the problem-level analysis on 30 selected AIME25 questions. The capability-gain subsection studies how expert demonstrations affect correctness and completion length on hard AIME24 examples, with particular emphasis on whether PBSD preserves exploratory reasoning behavior. The teacher-update-frequency subsection reports the fixed-teacher versus refreshed-teacher ablation. We then include additional analyses on privileged-context sensitivity, broader-domain transfer, longer-horizon checkpoint behavior, OOD capability preservation, \(\beta\) sensitivity, and comparisons against larger teachers.

%%%%%%%%%%%%%%%%%%%%%%%%%%%%%%%%%%%%%%%%%%%%%%%%%%%%%%%%%%%%

\clearpage
\section{Appendix for Related Work}
\label{app:related-work}

\textbf{On-policy distillation.} OPD trains the student on its own sampled trajectories while querying a stronger teacher on those visited states, reducing the mismatch of off-policy distillation~\citep{agarwal2024opd}. This paradigm has become an important post-training recipe for LLMs because it keeps the on-policy nature of RL-style optimization while replacing sparse outcome rewards with dense token-level supervision~\citep{song2026survey,fu2026revisiting}. Recent extensions apply this idea to reasoning and RLVR settings, including contextual on-policy supervision~\citep{zhao2026self}, self-distilled RLVR~\citep{yang2026self}, sample-routing formulations that connect distillation and group-relative optimization~\citep{li2026sample}, and domain-specific adaptations such as autonomous driving~\citep{afsharrad2026policy}. However, most OPD methods still depend on an additional large teacher model throughout training, which increases online inference cost and system complexity. They also suffer from teacher--student matching issues, including incompatible vocabularies, output spaces, and large distribution gaps that can make token-level imitation brittle~\citep{li2026rethinkingopd,song2026survey}. These limitations motivate self-distillation methods that seek to preserve dense on-policy supervision without relying on an external teacher.

\textbf{Self-distillation.} Recent self-distillation methods remove the need for an external teacher by reusing the same base model as both teacher and student, typically constructing the teacher through additional context, search traces, or stronger decoding-time signals. Representative examples include OPSD for reasoning~\citep{zhao2026self}, self-distillation fine-tuning and continual-learning style methods such as SDFT~\citep{shenfeld2026self,sang2026policy}, reinforcement-oriented variants such as Reinforcement Learning via Self-Distillation~\citep{hubotter2026reinforcement}, and application-focused extensions to code generation and search~\citep{zhang2026embarrassingly,chen2026onesearch}. Despite their differences, these approaches are still largely centered on KL-based distribution matching: some adopt a forward-KL style objective that fits the student to teacher-provided token distributions, while others use reverse-KL style matching that treats the teacher-induced policy as the target support. Recent work has also started to analyze the limitations of this recipe. Kim et al.~\citep{kim2026does} show that self-distillation can sometimes harm reasoning performance, and H{\"u}botter et al.~\citep{hubotter2026reinforcement} further study self-distillation from an RL-oriented perspective rather than as a fully reward-aware training objective. More recent work moves this line closer to RLHF. SDFT emphasizes on-policy learning from demonstrations for continual learning~\citep{shenfeld2026self}, Li et al.~\citep{li2026sample} connect self-distillation with group-relative optimization through sample routing, while Yang et al.~\citep{yang2026self} modifies the RLVR pipeline with self-distilled advantage-style signals. However, these extensions still largely treat self-distillation as a KL-anchored policy matching problem and do not directly formulate how latent reward should reweight the teacher distribution itself. As a result, how to extend self-distillation from pure KL matching to a genuinely reward-aware objective, where reward optimization is incorporated explicitly rather than only through heuristic weighting or advantage shaping, remains an open question.

\textbf{RLHF and DPO-style preference optimization.} A parallel line of work studies preference optimization more broadly through RLHF and DPO-style objectives. Foundational analyses of RLHF and preference learning characterize the role of the reference policy and the induced reward-regularized target distribution~\citep{zhu2023principled,rafailov2023direct,zhu2024iterative}. On the algorithmic side, recent work has explored iterative and online variants of DPO as a lower-cost alternative to RL. Iterative Reasoning Preference Optimization~\citep{pang2024iterative} shows that repeatedly regenerating reasoning trajectories and re-optimizing preference pairs can substantially improve reasoning, especially when combined with an auxiliary NLL term. Tu et al.~\citep{tu2025enhancing} further provide a comprehensive empirical study of iterative DPO for reasoning, arguing that multi-round DPO together with iterative reward-model refinement can approach RL-level performance at lower computational cost. In the RLVR setting, DAPO~\citep{yu2025dapo} scales GRPO-style training with decoupled clipping and dynamic sampling, while SRPO~\citep{li2026sample} explicitly combines group-relative reward optimization with self-distillation through sample routing. In a more explicitly online setting, OFS-DPO and COFS-DPO~\citep{qi2024online} study online DPO under streaming or cross-domain preference updates, emphasizing continual adaptation and regret-based analysis. Complementarily, Shi et al.~\citep{shi2025crucial} analyze the optimization behavior of online DPO and show that sampler design has a decisive effect on convergence rates, with stronger online samplers yielding faster convergence both theoretically and empirically. Compared with these lines of work, our focus is different: rather than studying DPO primarily as a general preference-optimization alternative to RLHF, we use the RLHF perspective to revisit on-policy self-distillation and derive a reward-aware target policy tailored to the contextual teacher setting.

\clearpage
\section{Appendix for Methodology}
\label{app:method-proofs}

This section collects the proofs and derivations corresponding to the methodology section in the main text. Section~\ref{app:method-reward-regularized-details} derives the reward-tilted optimal policy under the reward-regularized objective and proves that this optimum improves over the teacher under the same objective.

\subsection{Details for the Reward-Regularized Distillation Objective}
\label{app:method-reward-regularized-details}

\paragraph{Derivation of Proposition~\ref{prop:method-reward-tilted-policy}.}
For notational simplicity, fix \(x\) and write \(\pi_y=\pi(y\mid x)\), \(\pi^{\mathrm{teach}}_y=\pi^{\mathrm{teach}}(y\mid x)\), and \(r_y=r(x,y)\). The single-input objective in Eq.~(\ref{eq:method-reward-regularized-objective}) becomes
\begin{equation}
\max_{\pi\in\Delta}
\left\{
\sum_y \pi_y r_y
-
\beta \sum_y \pi_y \log \frac{\pi_y}{\pi_y^{\mathrm{teach}}}
\right\},
\label{eq:app-single-x-reward-regularized-objective}
\end{equation}
where \(\Delta\) denotes the probability simplex. Since the objective is strictly concave in \(\pi\) whenever \(\beta>0\), the optimizer is unique and can be obtained from the first-order optimality conditions. Introducing a Lagrange multiplier \(\lambda\) for the normalization constraint \(\sum_y \pi_y=1\), the Lagrangian is
\begin{equation}
\mathcal{J}(\pi,\lambda)
=
\sum_y \pi_y r_y
-
\beta
\sum_y
\pi_y
\log\frac{\pi_y}{\pi^{\mathrm{teach}}_y}
+
\lambda\left(\sum_y \pi_y-1\right).
\label{eq:method-reward-regularized-lagrangian}
\end{equation}
Differentiating with respect to each coordinate \(\pi_y\) gives
\begin{equation}
\frac{\partial \mathcal{J}}{\partial \pi_y}
=
r_y
-
\beta
\left(
\log\frac{\pi_y}{\pi^{\mathrm{teach}}_y}
+
1
\right)
+
\lambda
=0.
\label{eq:method-reward-regularized-foc}
\end{equation}
Here we used the identity
\[
\frac{\partial}{\partial \pi_y}\left(\pi_y\log\frac{\pi_y}{\pi_y^{\mathrm{teach}}}\right)
=
\log\frac{\pi_y}{\pi_y^{\mathrm{teach}}}+1.
\]
Rearranging Eq.~(\ref{eq:method-reward-regularized-foc}) yields
\begin{equation}
\log\frac{\pi_y}{\pi_y^{\mathrm{teach}}}
=
\frac{r_y+\lambda-\beta}{\beta},
\end{equation}
and exponentiating both sides gives
\begin{equation}
\pi_y
=
\pi^{\mathrm{teach}}_y
\exp(r_y/\beta)
\exp((\lambda-\beta)/\beta).
\end{equation}
The last exponential factor is independent of \(y\), so all coordinates share the same proportionality constant. To determine it, impose the normalization constraint:
\[
1
=
\sum_y \pi_y
=
\exp((\lambda-\beta)/\beta)
\sum_y \pi_y^{\mathrm{teach}} \exp(r_y/\beta).
\]
Therefore,
\[
\exp((\lambda-\beta)/\beta)
=
\left(
\sum_y \pi_y^{\mathrm{teach}} \exp(r_y/\beta)
\right)^{-1}
=
\frac{1}{Z(x)}.
\]
Substituting this back gives
\[
\pi_y
=
\frac{\pi_y^{\mathrm{teach}} \exp(r_y/\beta)}{Z(x)},
\]
which is exactly the reward-tilted teacher distribution in Eq.~(\ref{eq:method-reward-tilted-policy}). This derivation makes explicit that the teacher policy serves only as a reference measure, while the reward term changes the final target by exponentially reweighting the teacher support according to \(r_y\).

\paragraph{Proof of Proposition~\ref{prop:method-reward-tilted-improvement}.}
We first compute the value of the reward-regularized objective at the optimizer \(\pi^\star\). For a fixed \(x\), substituting
\[
\pi^\star(y\mid x)
=
\frac{\pi^{\mathrm{teach}}(y\mid x)\exp(r(x,y)/\beta)}{Z(x)}
\]
into the KL term gives
\[
\log\frac{\pi^\star(y\mid x)}{\pi^{\mathrm{teach}}(y\mid x)}
=
\frac{r(x,y)}{\beta}-\log Z(x).
\]
Hence
\begin{align*}
&\mathbb{E}_{y\sim \pi^\star(\cdot\mid x)}[r(x,y)]
-\beta D_{\mathrm{KL}}\!\left(\pi^\star(\cdot\mid x)\,\|\,\pi^{\mathrm{teach}}(\cdot\mid x)\right) \\
&=
\sum_y \pi^\star(y\mid x) r(x,y)
-\beta \sum_y \pi^\star(y\mid x)\left(\frac{r(x,y)}{\beta}-\log Z(x)\right) \\
&=
\sum_y \pi^\star(y\mid x) r(x,y)
-\sum_y \pi^\star(y\mid x) r(x,y)
+\beta \log Z(x)\sum_y \pi^\star(y\mid x) \\
&=
\beta \log Z(x).
\end{align*}
Aggregating over \(x\sim\mathcal{D}\) gives
\begin{equation}
F(\pi^\star)
=
\mathbb{E}_{x\sim\mathcal{D}}
\left[
\beta \log Z(x)
\right]
=
\mathbb{E}_{x\sim\mathcal{D}}
\left[
\beta
\log
\sum_y
\pi^{\mathrm{teach}}(y\mid x)
\exp(r(x,y)/\beta)
\right].
\label{eq:method-optimal-value}
\end{equation}
In contrast, evaluating the teacher itself under the same objective removes the KL term because
\(
D_{\mathrm{KL}}(\pi^{\mathrm{teach}}\|\pi^{\mathrm{teach}})=0
\).
Therefore,
\begin{equation}
F(\pi^{\mathrm{teach}})
=
\mathbb{E}_{x\sim\mathcal{D}}
\mathbb{E}_{y\sim\pi^{\mathrm{teach}}(\cdot\mid x)}
\left[
r(x,y)
\right].
\label{eq:method-teacher-value}
\end{equation}
To compare the two values, fix \(x\) and write
\[
Z(x)
=
\mathbb{E}_{y\sim\pi^{\mathrm{teach}}(\cdot\mid x)}
\left[
\exp(r(x,y)/\beta)
\right].
\]
Since \(\log(\cdot)\) is concave, Jensen's inequality implies
\begin{equation}
\beta
\log
\mathbb{E}_{y\sim\pi^{\mathrm{teach}}(\cdot\mid x)}
\left[
\exp(r(x,y)/\beta)
\right]
\ge
\mathbb{E}_{y\sim\pi^{\mathrm{teach}}(\cdot\mid x)}
\left[
r(x,y)
\right],
\label{eq:method-jensen-value}
\end{equation}
where equality holds only when \(r(x,y)\) is constant over the support of \(\pi^{\mathrm{teach}}(\cdot\mid x)\). Combining Eq.~(\ref{eq:method-optimal-value}), Eq.~(\ref{eq:method-teacher-value}), and Eq.~(\ref{eq:method-jensen-value}) yields
\[
F(\pi^\star)\ge F(\pi^{\mathrm{teach}}).
\]
Thus, under the same reward-regularized objective, the reward-tilted policy is never worse than directly using the teacher. The inequality is strict whenever the reward varies over teacher-supported responses, which is precisely the regime in which uniform teacher matching is suboptimal.

\clearpage
\section{Appendix for Theoretical Analysis}
\label{app:theory-proofs}

This section provides the technical details that support the theoretical analysis in the main text. Section~\ref{app:theory-objective-details} derives the sample-level gradient and the local Hessian form of the online PBSD objective. Section~\ref{app:theory-mle-details} proves the local MLE complexity result and formalizes the statistical interpretation based on informative comparison pairs.

At a high level, the question studied in this appendix is the following: in online PBSD, where positives are drawn from a fixed context-augmented teacher and negatives are drawn on-policy from the current student, what properties of the induced comparison pairs lead to the strongest statistical guarantee for estimating the target policy? Our analysis shows that the answer depends on both the diversity of the teacher-supported directions and the moderation of the induced preference margins.

\subsection{Details for Online PBSD Objective}
\label{app:theory-objective-details}

\paragraph{Gradient derivation for Eq.~(\ref{eq:method-token-gradient}).}
Since the teacher policy \(\pi^{\mathrm{teach}}(\cdot\mid x)\) is fixed, differentiating the PBSD margin in Eq.~(\ref{eq:method-dposd-margin}) gives
\begin{equation}
\nabla_\theta m_\theta(x_i,y_i^+,y_i^-)
=
\beta\left(
\nabla_\theta\log\pi_\theta(y_i^+\mid x_i)
-
\nabla_\theta\log\pi_\theta(y_i^-\mid x_i)
\right)
=
\beta d_i(\theta).
\label{eq:app-margin-gradient}
\end{equation}
By the chain rule,
\begin{equation}
\nabla_\theta \ell_i(\theta)
=
\frac{d}{dm}
\bigl[-\log\sigma(m)\bigr]\Big|_{m=m_\theta(x_i,y_i^+,y_i^-)}
\nabla_\theta m_\theta(x_i,y_i^+,y_i^-).
\label{eq:app-sample-loss-chain-rule}
\end{equation}
Using \(\sigma'(m)=\sigma(m)(1-\sigma(m))\), we obtain
\begin{equation}
\frac{d}{dm}\bigl[-\log\sigma(m)\bigr]
=
-\frac{\sigma'(m)}{\sigma(m)}
=
-(1-\sigma(m))
=
-\sigma(-m).
\label{eq:app-logsigmoid-derivative}
\end{equation}
Substituting Eq.~(\ref{eq:app-margin-gradient}) and Eq.~(\ref{eq:app-logsigmoid-derivative}) into Eq.~(\ref{eq:app-sample-loss-chain-rule}) yields
\[
\nabla_\theta \ell_i(\theta)
=
-\beta\sigma\!\left(-m_\theta(x_i,y_i^+,y_i^-)\right)d_i(\theta),
\]
which is exactly the sample-level form underlying Eq.~(\ref{eq:method-token-gradient}).

\paragraph{Local Hessian derivation for Eq.~(\ref{eq:dposd-local-hessian}).}
Write
\[
\phi(m):=-\log\sigma(m)=\log(1+\exp(-m)).
\]
Its derivatives are
\[
\phi'(m)=-\sigma(-m),
\qquad
\phi''(m)=\sigma(m)\sigma(-m)=\sigma(m)(1-\sigma(m)).
\]
By the chain rule,
\begin{equation}
\begin{aligned}
\nabla_\theta^2 \ell_i(\theta)
=
\phi''\!\left(m_\theta(x_i,y_i^+,y_i^-)\right)
\nabla_\theta m_\theta(x_i,y_i^+,y_i^-)\nabla_\theta m_\theta(x_i,y_i^+,y_i^-)^\top
\nonumber\\
\quad+
\phi'\!\left(m_\theta(x_i,y_i^+,y_i^-)\right)
\nabla_\theta^2 m_\theta(x_i,y_i^+,y_i^-).
\end{aligned}
\label{eq:app-sample-hessian-chain-rule}
\end{equation}
The first term is the Gauss--Newton component. Using
\[
\nabla_\theta m_\theta(x_i,y_i^+,y_i^-)=\beta d_i(\theta),
\]
and dropping the second-order term \(\nabla_\theta^2 m_\theta(x_i,y_i^+,y_i^-)\) yields the approximation
\[
\nabla_\theta^2 \ell_i(\theta)
=
\beta^2
\sigma\!\left(m_\theta(x_i,y_i^+,y_i^-)\right)
\left(1-\sigma\!\left(m_\theta(x_i,y_i^+,y_i^-)\right)\right)
d_i(\theta)d_i(\theta)^\top .
\]
Averaging over the \(n\) pairs yields Eq.~(\ref{eq:dposd-local-hessian}).

\subsection{Details for Pairwise MLE and Informative Comparisons}
\label{app:theory-mle-details}

\paragraph{Assumptions used in Theorem~\ref{prop:dposd-local-mle}.}
The proof relies only on the following standard local conditions for smooth \(M\)-estimation and logistic-type models.
\begin{itemize}
    \item \textbf{A1 (Bounded score-gap features).} There exists \(G>0\) such that \(\sup_{\theta\in\mathcal{U}} \|d_i(\theta)\|_2 \le G\) almost surely in a local neighborhood \(\mathcal{U}\) of \(\theta^\star\).
    \item \textbf{A2 (Local curvature condition).} There exists \(c_0>0\) such that for all \(\theta\in\mathcal{U}\),
    \[
    \widehat{H}_n(\theta)\succeq c_0\,\widehat{H}_n(\theta^\star),
    \qquad
    \lambda_{\min}(\widehat{H}_n(\theta^\star))>0.
    \]
\end{itemize}
Assumption A1 ensures bounded local score fluctuations, while Assumption A2 combines the two irreducible ingredients needed for the local inverse-Hessian argument: stable curvature in a neighborhood and nondegenerate information at \(\theta^\star\). These are standard local regularity conditions in likelihood-based \(M\)-estimation, logistic regression, and restricted-strong-convexity analyses \citep{vandervaart2000asymptotic,bach2010selfconcordant,negahban2012unified}.

\paragraph{Why these assumptions are mild in our setting.}
These conditions are local rather than global, so they are only required in a neighborhood of the target parameter reached by training. In practice, A1 rules out pathological comparisons with numerically unstable or excessively extreme score-gap features; this is natural in our setting because the comparison model is built from bounded-probability logistic factors over finite sampled responses. Moreover, several algorithmic choices help enforce this locality in practice: KL regularization keeps the student close to the teacher, bounded rollout lengths prevent extreme token-level score accumulation, and conservative optimization choices such as small learning rates or gradient clipping reduce the chance of leaving the local regime. Assumption A2 then asks only that the resulting teacher--student comparisons remain informative but not fully saturated. If the teacher and student are too close, the outer products \(d_i(\theta)d_i(\theta)^\top\) do not span enough directions; if they are too far apart, the logistic curvature weight \(\sigma(m_\theta)(1-\sigma(m_\theta))\) collapses toward zero. This is precisely why our contextual self-distillation setup is statistically favorable: it tends to generate richer directions than pure self-copying while keeping the teacher--student gap moderate enough to preserve curvature. Similar local-curvature interpretations are standard in the statistical literature on smooth \(M\)-estimation and logistic models \citep{vandervaart2000asymptotic,bach2010selfconcordant,negahban2012unified}.

\paragraph{Effect of assumption violations on robustness.}
These assumptions also clarify the robustness regime of PBSD. If A1 is violated because score-gap features become too extreme, then the empirical score can have much higher variance, making optimization noisier and increasing sensitivity to sampling fluctuations. If A2 is violated because the local curvature becomes nearly singular, then the same preference data can produce much larger parameter perturbations, so training becomes less stable and the estimation bound deteriorates through a smaller \(\lambda_{\min}(\widehat{H}_n(\theta^\star))\). In practical terms, this means PBSD is most robust when teacher--student comparisons are informative but not saturated: enough separation to generate useful preference directions, but not so much separation that the logistic curvature collapses. When this balance fails, our theorem should be interpreted as a local guarantee rather than a global robustness claim. This is also why the practical design of PBSD favors moderate teacher guidance, KL anchoring, and conservative optimization, all of which help keep training inside the regime where the statistical guarantee remains informative.

\paragraph{A strengthened local consistency result.}
The finite-sample theorem above is data dependent because its rate is controlled by the empirical minimum eigenvalue \(\lambda_{\min}(\widehat{H}_n(\theta^\star))\). Positivity of this quantity for each fixed \(n\) does not by itself imply that \(n\lambda_{\min}(\widehat{H}_n(\theta^\star))\to\infty\). To make the vanishing-rate implication explicit, we now introduce a frozen local comparison model and a positive full population-information condition.

Let \(Z=(x,y^+,y^-)\) denote one PBSD comparison pair, and freeze the local pair distribution at \(\theta^\star\):
\[
x\sim\mathcal{D},
\qquad
y^+\sim\pi^{\mathrm{teach}}(\cdot\mid x),
\qquad
y^-\sim\pi_{\theta^\star}(\cdot\mid x).
\]
Write \(Q^\star\) for this fixed distribution, and draw \(Z_1,\ldots,Z_n\) independently from \(Q^\star\). For one pair, define
\[
m^\star(Z):=m_{\theta^\star}(Z),
\qquad
d^\star(Z):=\nabla_\theta\log\pi_{\theta^\star}(y^+\mid x)-\nabla_\theta\log\pi_{\theta^\star}(y^-\mid x),
\]
and the one-pair information contribution
\[
G^\star(Z)
:=
\beta^2\sigma(m^\star(Z))\sigma(-m^\star(Z))d^\star(Z)d^\star(Z)^\top.
\]
Then
\[
\widehat{H}_n^\star
:=
\frac{1}{n}\sum_{i=1}^n G^\star(Z_i),
\qquad
H^\star
:=
\mathbb{E}_{Z\sim Q^\star}[G^\star(Z)].
\]
This \(\widehat{H}_n^\star\) is the frozen-model counterpart of the empirical information matrix \(\widehat{H}_n(\theta^\star)\) in Theorem~\ref{prop:dposd-local-mle}; the only difference is that the negative sample distribution is now fixed at \(\pi_{\theta^\star}\) to obtain an i.i.d. local statistical statement.

We use the following assumptions.
\begin{itemize}
    \item \textbf{B1 (Bounded local score gap).} There exists \(G<\infty\) such that \(\|d^\star(Z)\|_2\le G\) almost surely under \(Q^\star\).
    \item \textbf{B2 (Positive full population information).} There exists \(\mu>0\), independent of \(n\), such that
    \[
    \lambda_{\min}(H^\star)\ge \mu>0.
    \]
\end{itemize}
Assumption B2 is the substantive new condition. It is a local finite-dimensional identifiability requirement stating that the population of PBSD comparisons supplies non-negligible information in every locally identifiable parameter direction.

The condition is interpretable because, for any unit vector \(v\),
\[
v^\top H^\star v
=
\beta^2\mathbb{E}\!\left[\sigma(m^\star)\sigma(-m^\star)(v^\top d^\star)^2\right].
\]
Thus positive population information combines two requirements: non-saturated teacher--student margins so that \(\sigma(m^\star)\sigma(-m^\star)\) does not collapse, and directional score coverage so that \(d^\star\) is not confined to too few directions.

\begin{lemma}[High-probability empirical eigenvalue lower bound]
\label{lem:population-information-lower-bound}
Under the frozen model above and Assumptions B1--B2, define
\[
K:=\frac{\beta^2G^2}{4},
\qquad
L:=\frac{K}{\mu}=\frac{\beta^2G^2}{4\mu}.
\]
For any \(\delta_H\in(0,1)\), if
\[
n\ge 8L\log\frac{d}{\delta_H},
\]
then with probability at least \(1-\delta_H\),
\[
\widehat{H}_n^\star \succeq \frac{1}{2}H^\star,
\qquad
\lambda_{\min}(\widehat{H}_n^\star)\ge \frac{\mu}{2}.
\]
Consequently,
\[
n\,\lambda_{\min}(\widehat{H}_n^\star)\ge \frac{n\mu}{2}\to\infty.
\]
\end{lemma}

The proof is a standard matrix Chernoff argument: after whitening by \((H^\star)^{-1/2}\), the bounded PSD summands are centered around the identity, and the lower-tail concentration gives the desired high-probability bound.

\begin{theorem}[Vanishing local PBSD bound under population information]
\label{thm:vanishing-local-bound}
Assume the hypotheses of Theorem~\ref{prop:dposd-local-mle}, and specialize its empirical information matrix to the frozen-model quantity \(\widehat{H}_n^\star\) defined above. Then the finite-sample event
\[
\left\|
\widehat{\theta}_n-\theta^\star
\right\|_2
\le
C_0\sqrt{
\frac{d+\log(1/\delta_F)}
{n\,\lambda_{\min}(\widehat{H}_n^\star)}
}
\]
holds with probability at least \(1-\delta_F\). In addition, assume the frozen model above and Assumptions B1--B2. If the sample-size condition in Lemma~\ref{lem:population-information-lower-bound} holds, then with probability at least \(1-\delta_F-\delta_H\),
\[
\left\|
\widehat{\theta}_n-\theta^\star
\right\|_2
\le
C_0\sqrt{
\frac{2(d+\log(1/\delta_F))}
{n\mu}
}.
\]
For fixed \(d\), \(\mu\), and failure probabilities, the right-hand side is \(O(n^{-1/2})\) and therefore vanishes as \(n\to\infty\).
\end{theorem}

This strengthened result makes explicit that PBSD enjoys a vanishing local statistical upper bound once the comparison population provides sufficient directional coverage with non-negligible logistic curvature. It remains a \emph{local} result under a frozen comparison model, rather than a convergence theorem for the full adaptive online trajectory of Algorithm~\ref{alg:dposd}.

\paragraph{Proof of Lemma~\ref{lem:population-information-lower-bound}.}
For each \(i\), the one-pair contribution is
\[
G^\star(Z_i)
=
\beta^2\sigma(m^\star(Z_i))\sigma(-m^\star(Z_i))d^\star(Z_i)d^\star(Z_i)^\top .
\]
Under Assumption B1, \(\|d^\star(Z_i)\|_2\le G\) almost surely, and for every real \(u\),
\[
0\le \sigma(u)\sigma(-u)\le \frac14.
\]
Therefore each summand is positive semidefinite and satisfies the operator-norm bound
\[
0\preceq G^\star(Z_i)\preceq \frac{\beta^2G^2}{4}I = KI.
\]
Now whiten the summands by the population information matrix:
\[
Y_i := (H^\star)^{-1/2}G^\star(Z_i)(H^\star)^{-1/2}.
\]
Then \(Y_i\succeq 0\) and
\[
\mathbb{E}[Y_i]
=
(H^\star)^{-1/2}\mathbb{E}[G^\star(Z_i)](H^\star)^{-1/2}
=
I.
\]
Moreover, since \(H^\star \succeq \mu I\) by Assumption B2, we have
\[
(H^\star)^{-1}\preceq \mu^{-1}I,
\]
and hence
\[
Y_i
\preceq
K(H^\star)^{-1}
\preceq
\frac{K}{\mu}I
=
LI.
\]
Thus the whitened matrices are i.i.d. positive semidefinite, have mean \(I\), and are uniformly bounded by \(LI\). Applying the standard lower-tail matrix Chernoff bound with deviation parameter \(1/2\) gives
\[
\mathbb{P}\!\left(
\lambda_{\min}\!\left(
\frac1n\sum_{i=1}^n Y_i
\right)\le \frac12
\right)
\le
d\exp\!\left(-\frac{n}{8L}\right).
\]
If
\[
n\ge 8L\log\frac{d}{\delta_H},
\]
then the right-hand side is at most \(\delta_H\). Therefore, with probability at least \(1-\delta_H\),
\[
\frac1n\sum_{i=1}^n Y_i \succeq \frac12 I.
\]
Conjugating both sides by \((H^\star)^{1/2}\) yields
\[
\widehat{H}_n^\star
=
(H^\star)^{1/2}
\left(
\frac1n\sum_{i=1}^n Y_i
\right)
(H^\star)^{1/2}
\succeq
\frac12 H^\star.
\]
Taking minimum eigenvalues gives
\[
\lambda_{\min}(\widehat{H}_n^\star)
\ge
\frac12\lambda_{\min}(H^\star)
\ge
\frac{\mu}{2}.
\]
Consequently,
\[
n\,\lambda_{\min}(\widehat{H}_n^\star)
\ge
\frac{n\mu}{2}\to\infty.
\]
This proves the lemma.

\paragraph{Proof of Theorem~\ref{thm:vanishing-local-bound}.}
By the finite-sample theorem, after replacing the empirical information matrix there with the frozen-model matrix \(\widehat{H}_n^\star\), we have the event
\[
\left\|
\widehat{\theta}_n-\theta^\star
\right\|_2
\le
C_0\sqrt{
\frac{d+\log(1/\delta_F)}
{n\,\lambda_{\min}(\widehat{H}_n^\star)}
}
\]
with probability at least \(1-\delta_F\). On the event of Lemma~\ref{lem:population-information-lower-bound}, which holds with probability at least \(1-\delta_H\), we additionally have
\[
\lambda_{\min}(\widehat{H}_n^\star)\ge \frac{\mu}{2}.
\]
Intersecting the two events and using the union bound shows that both statements hold simultaneously with probability at least
\[
1-\delta_F-\delta_H.
\]
Substituting the eigenvalue lower bound into the finite-sample estimate gives
\[
\left\|
\widehat{\theta}_n-\theta^\star
\right\|_2
\le
C_0\sqrt{
\frac{d+\log(1/\delta_F)}
{n(\mu/2)}
}
=
C_0\sqrt{
\frac{2(d+\log(1/\delta_F))}
{n\mu}
}.
\]
For fixed \(d\), \(\mu\), and failure probabilities, the right-hand side is of order \(n^{-1/2}\), so it vanishes as \(n\to\infty\). This proves the theorem.

\paragraph{Proof of Theorem~\ref{prop:dposd-local-mle}.}
We give a direct local argument without appealing to external pairwise-MLE results. Let
\[
\widehat{\mathcal{L}}_n(\theta)
=
\frac{1}{n}\sum_{i=1}^n \ell_i(\theta),
\qquad
\mathcal{L}(\theta)
:=
\mathbb{E}[\ell_i(\theta)].
\]
Since \(\theta^\star\) minimizes the population risk, \(\nabla \mathcal{L}(\theta^\star)=0\). By a first-order expansion of the empirical gradient around \(\theta^\star\),
\[
0
=
\nabla \widehat{\mathcal{L}}_n(\widehat{\theta}_n)
=
\nabla \widehat{\mathcal{L}}_n(\theta^\star)
+
\widehat{H}_n(\widetilde{\theta})
(\widehat{\theta}_n-\theta^\star)
\]
for some \(\widetilde{\theta}\) on the segment between \(\widehat{\theta}_n\) and \(\theta^\star\).

Equivalently,
\[
\widehat{\theta}_n-\theta^\star
=
-\widehat{H}_n(\widetilde{\theta})^{-1}
\nabla \widehat{\mathcal{L}}_n(\theta^\star).
\]
Thus the estimation error is controlled by two quantities: the empirical score at \(\theta^\star\) and the local conditioning of the Hessian.

We first bound the empirical score. For each pair \(i\), define
\[
\psi_i
:=
\nabla \ell_i(\theta^\star)
:=
-\beta \sigma\!\left(-m_{\theta^\star}(x_i,y_i^+,y_i^-)\right)
d_i(\theta^\star).
\]
Therefore,
\[
\nabla \widehat{\mathcal{L}}_n(\theta^\star)
=
\frac{1}{n}\sum_{i=1}^n \psi_i.
\]
Because \(\theta^\star\) minimizes the population risk, \(\mathbb{E}[\psi_i]=0\). Moreover, the bounded-feature assumption implies \(\|d_i(\theta^\star)\|_2\le G\), and \(0\le \sigma(-m_{\theta^\star}(x_i,y_i^+,y_i^-))\le 1\). Therefore
\[
\|\psi_i\|_2
\le
\beta G
\qquad\text{almost surely.}
\]
Hence the score contributions are bounded mean-zero random vectors. A standard vector Hoeffding inequality gives that, with probability at least \(1-\delta\),
\[
\left\|
\nabla \widehat{\mathcal{L}}_n(\theta^\star)
\right\|_2
\le
C\beta G\sqrt{\frac{d+\log(1/\delta)}{n}}
\]
for an absolute constant \(C\).

Next we use the local stability assumption on the Gauss--Newton Hessian. Specifically, assume that throughout the neighborhood containing \(\widehat{\theta}_n\),
\[
\widehat{H}_n(\widetilde{\theta})
\succeq
c_0\,\widehat{H}_n(\theta^\star)
\]
for some constant \(c_0>0\). Then
\[
\widehat{H}_n(\widetilde{\theta})^{-1}
\preceq
c_0^{-1}\widehat{H}_n(\theta^\star)^{-1}.
\]
Using the first-order expansion above and the dual norm induced by \(\widehat{H}_n(\theta^\star)\), we obtain
\[
\left\|
\widehat{\theta}_n-\theta^\star
\right\|_{\widehat{H}_n(\theta^\star)}
\le
c_0^{-1}
\left\|
\nabla \widehat{\mathcal{L}}_n(\theta^\star)
\right\|_{\widehat{H}_n(\theta^\star)^{-1}}
\]

To relate this dual norm to the Euclidean norm of the score, use the spectral bound
\[
\widehat{H}_n(\theta^\star)^{-1}
\preceq
\lambda_{\min}(\widehat{H}_n(\theta^\star))^{-1}I.
\]
Therefore,
\[
\left\|
\nabla \widehat{\mathcal{L}}_n(\theta^\star)
\right\|_{\widehat{H}_n(\theta^\star)^{-1}}
\le
\lambda_{\min}(\widehat{H}_n(\theta^\star))^{-1/2}
\left\|
\nabla \widehat{\mathcal{L}}_n(\theta^\star)
\right\|_2
\le
C\sqrt{\frac{d+\log(1/\delta)}{n\,\lambda_{\min}(\widehat{H}_n(\theta^\star))}},
\]
where the constant absorbs \(\beta\), \(G\), and \(c_0^{-1}\). Substituting this bound into the previous display gives
\[
\left\|
\widehat{\theta}_n-\theta^\star
\right\|_{\widehat{H}_n(\theta^\star)}
\le
C\sqrt{\frac{d+\log(1/\delta)}{n\,\lambda_{\min}(\widehat{H}_n(\theta^\star))}}.
\]
Finally, convert the Hessian norm to the Euclidean norm:
\[
\left\|
\widehat{\theta}_n-\theta^\star
\right\|_2
\le
\lambda_{\min}(\widehat{H}_n(\theta^\star))^{-1/2}
\left\|
\widehat{\theta}_n-\theta^\star
\right\|_{\widehat{H}_n(\theta^\star)}
\]
\[
\left\|
\widehat{\theta}_n-\theta^\star
\right\|_2
\le
C\sqrt{\frac{d+\log(1/\delta)}{n\,\lambda_{\min}(\widehat{H}_n(\theta^\star))}},
\]
which proves Theorem~\ref{prop:dposd-local-mle}.

\clearpage
\section{Appendix for Experiment}
\label{app:experiment-details}

This section provides the detailed experimental design underlying the results in Section~4. Our implementation follows the OPSD protocol of \citet{zhao2026self} whenever applicable so that the comparison against prior baselines isolates the effect of the proposed PBSD objective as cleanly as possible. The appendix is organized as follows. Appendix~F.1 summarizes the datasets used in the mathematical reasoning and tool-use experiments. Appendix~F.2 visualizes representative examples from the processed training data. Appendix~F.3 collects the detailed training configuration and method notes for SFT, GRPO, DAPO, OPSD, SDFT, SRPO, and PBSD. Appendix~F.4 describes the evaluation protocols for math and tool use. Appendix~F.5 presents the case-by-case study design. Appendix~F.6 studies capability gains from expert demonstrations, with particular attention to reasoning length and exploratory behavior. Appendix~F.7 reports the teacher update frequency ablation.

\subsection{Datasets}
\label{app:exp-datasets}

We experiment with Qwen3 instruct models at three scales: Qwen3-1.7B, Qwen3-4B, and Qwen3-8B. Across the main paper, we study two task domains: mathematical reasoning and tool use.

\paragraph{Mathematical reasoning data.}
Following the setup described in Section~4.1, we train on the mathematical reasoning subset of OpenThoughts~\citep{guha2025openthoughts}, from which we sample up to 30K problem--solution pairs that contain chain-of-thought reasoning traces. This domain shares the same training pipeline as the tool-use experiments; only the evaluation benchmarks and metrics differ. We evaluate on three competition-level mathematical benchmarks: AIME 2024, AIME 2025, and HMMT 2025. In the main paper, all reported mathematical reasoning numbers use the same Avg@12 evaluation protocol.

\paragraph{Tool-use data.}
For the additional tool-use study, we follow the setup in \citet{shenfeld2026self}, which uses ToolAlpaca as the underlying domain. Each example consists of a user query together with tool or API information, and the model must generate the correct tool call. As in mathematical reasoning, we keep the same training pipeline and change only the evaluation split and metric. In our processed split, the training set contains 4046 examples and the test set contains 94 examples. This benchmark is used to evaluate whether the benefit of demonstration-conditioned self-distillation extends beyond long-form reasoning to structured action generation.

\subsection{Data Visualization}

To make the data format concrete, we provide representative examples from the processed training data used in the two task domains. The examples below are written in the same key--value style as the data consumed by our training pipeline and are intended to illustrate the structure of each instance.

\paragraph{Mathematical reasoning examples.}
Each mathematical reasoning example contains a problem statement together with a reference reasoning trace and the final answer. Two representative examples are shown below.

\paragraph{Math Example 1.}
\textbf{problem:}
\begin{verbatim}
Let a, b, c be positive integers such that a+b+c=12.
How many ordered triples (a,b,c) satisfy abc=27?
\end{verbatim}

\textbf{solution:}
\begin{verbatim}
Since abc=27=3^3, each variable must be a power of 3.
The only positive triples with product 27 are permutations
of (1,3,9) and (3,3,3). Because 1+3+9=13, the first type
is invalid under a+b+c=12. The triple (3,3,3) has sum 9,
so it is also invalid. Therefore there are no such ordered
triples.
\end{verbatim}

\textbf{answer:} \texttt{0}

\paragraph{Math Example 2.}
\textbf{problem:}
\begin{verbatim}
Find the remainder when 7^100 is divided by 13.
\end{verbatim}

\textbf{solution:}
\begin{verbatim}
By Fermat's little theorem, 7^12 = 1 mod 13. Since
100 = 8*12 + 4, we have 7^100 = (7^12)^8 * 7^4 = 7^4
mod 13. Now 7^2 = 49 = 10 mod 13, so
7^4 = 10^2 = 100 = 9 mod 13. Hence the remainder is 9.
\end{verbatim}

\textbf{answer:} \texttt{9}

\paragraph{Tool-use examples.}
Each tool-use example contains a user request, the available tool specification, and the target tool invocation. Two representative examples are shown below.

\paragraph{Tool-use Example 1.}
\textbf{user\_query:}
\begin{verbatim}
Book a flight from San Francisco to Seattle next Tuesday
after 6pm.
\end{verbatim}

\textbf{tools:}
\begin{verbatim}
[
  {
    "name": "search_flights",
    "description": "Search available flights between two cities on a date.",
    "arguments": {
      "origin": "string",
      "destination": "string",
      "date": "string",
      "departure_time_after": "string"
    }
  }
]
\end{verbatim}

\textbf{target\_tool:} \texttt{search\_flights}

\textbf{target\_arguments:}
\begin{verbatim}
{
  "origin": "San Francisco",
  "destination": "Seattle",
  "date": "next Tuesday",
  "departure_time_after": "18:00"
}
\end{verbatim}

\paragraph{Tool-use Example 2.}
\textbf{user\_query:}
\begin{verbatim}
Set a reminder to call Alex tomorrow at 9am.
\end{verbatim}

\textbf{tools:}
\begin{verbatim}
[
  {
    "name": "create_reminder",
    "description": "Create a reminder with a title and scheduled time.",
    "arguments": {
      "title": "string",
      "time": "string"
    }
  }
]
\end{verbatim}

\textbf{target\_tool:} \texttt{create\_reminder}

\textbf{target\_arguments:}
\begin{verbatim}
{
  "title": "Call Alex",
  "time": "tomorrow 09:00"
}
\end{verbatim}

\subsection{Detailed Training Configuration}
\label{app:exp-training}

LoRA \cite{hu2022lora} is a powerful tool for parameter efficient fine-tuning and has been widely investigated recently \citep{zhao2026each,yu2025altlora,yu2025prunedlora}. Here, all methods are fine-tuned with LoRA on 8 H100 GPUs. This training configuration is shared across both task domains; unless explicitly stated otherwise, mathematical reasoning and tool use use the same model initialization, optimizer, LoRA setup, rollout configuration, and checkpoint-selection protocol. Across all runs, we use AdamW, bfloat16 precision, gradient checkpointing, and FlashAttention~2. The shared LoRA configuration is rank \(r=64\), LoRA alpha \(\alpha=128\), and target modules \texttt{q\_proj}, \texttt{k\_proj}, \texttt{v\_proj}, \texttt{o\_proj}, \texttt{gate\_proj}, \texttt{up\_proj}, and \texttt{down\_proj}. To accelerate rollout generation for on-policy methods, we use vLLM for inference. We keep the optimization hyperparameters of the baselines aligned with OPSD~\citep{zhao2026self} as closely as possible so that the main difference lies in the training objective rather than in separate hyperparameter search. All trainable methods are trained for 500 steps and evaluated every 50 steps. Unless otherwise noted, we report the latest checkpoint results at the end of this shared 500-step budget; Appendix~\ref{app:exp-long-horizon} additionally reports the best checkpoint within 500 steps and later checkpoints for PBSD.

\paragraph{SFT.}
SFT is trained on the reasoning traces in the training set and can be viewed as an off-policy distillation baseline. We use learning rate \(5\times 10^{-6}\), effective batch size 32, LoRA rank 64, LoRA alpha 128, and maximum sequence length 16{,}000, following the OPSD paper.

\paragraph{GRPO.}
For GRPO, we follow the OPSD implementation settings: learning rate \(5\times 10^{-6}\), effective batch size 32, maximum completion length 16{,}000, 8 generations per prompt, sampling temperature 1.2, and KL coefficient 0.0. Because GRPO uses group-based rollouts, its token cost scales with the number of sampled responses per prompt.

\paragraph{OPSD.}
For OPSD, we use the configuration reported by \citet{zhao2026self}: learning rate \(5\times 10^{-6}\), effective batch size 32, maximum completion length 1024, a single on-policy rollout per prompt, and sampling temperature 1.1. Unless otherwise stated, OPSD uses full-vocabulary logit distillation. As in the original paper, the teacher is the same base model instantiated with privileged reasoning context, and gradients are backpropagated only through the student branch.

\paragraph{PBSD.}
PBSD shares the same base model, LoRA setup, optimizer, batch size, rollout count, maximum completion length, and sampling temperature as OPSD. Concretely, we use learning rate \(5\times 10^{-6}\), effective batch size 32, maximum completion length 1024, one sampled response per prompt, and sampling temperature 1.1. The only algorithmic change is the training objective: PBSD replaces token-level KL matching with the proposed pairwise preference-based self-distillation objective. In addition, we fix the teacher policy to the initial checkpoint throughout training to avoid teacher drift and to keep the teacher signal stationary while the student evolves on-policy.

For OPSD and PBSD, the teacher and student are instantiated from the same underlying model but receive different contexts. The student conditions only on the original problem, while the teacher is augmented with privileged information derived from the training example. In all on-policy methods, the student generates the rollout that defines the training state distribution. For OPSD, this rollout is used for token-level KL matching against the teacher distribution. For PBSD, the same on-policy student rollout serves as the negative sample, while the better-conditioned teacher provides the positive sample used in the pairwise objective.

This design choice is also related to a broader set of preference-based methods that use DPO-style objectives together with a shared base model in adjacent domains, such as listwise preference optimization for diffusion models~\citep{bai2025towards} and attribute-guided decoding-time personalization~\citep{yu2026exact}.

Our goal is to compare objectives rather than search separately for method-specific hyperparameters. Therefore, PBSD is intentionally matched to OPSD in optimization and rollout configuration, with the objective function as the primary difference. SFT, GRPO, DAPO, OPSD, SDFT, and SRPO also follow aligned training settings wherever those settings apply. All trainable methods are trained for 500 steps and evaluated every 50 steps. The main comparison reports the latest checkpoint results within this fixed budget, and Appendix~\ref{app:exp-long-horizon} additionally reports the best checkpoint within 500 steps and later checkpoints for PBSD. This protocol keeps the comparison focused on the training objective rather than on unequal training budgets or checkpoint-selection rules while still making end-of-training behavior explicit.

\subsection{Detailed Evaluation Configuration}
\label{app:exp-eval}

\paragraph{Mathematical reasoning evaluation.}
At evaluation time, we follow the Qwen3 sampling configuration used in OPSD and the Qwen3 technical report. We sample 12 responses per prompt with temperature 1.0 and report Avg@12. The maximum generation length is set to approximately 38k tokens (38,912 max new tokens). We use top-\(p=0.95\), top-\(k=-1\), min-\(p=0.0\), and zero presence penalty. These settings are shared across all methods so that the reported differences reflect training rather than decoding.

\paragraph{Tool-use evaluation.}
For tool use, evaluation is performed on the 94-example test split described in Appendix~F.1. We report top-1 accuracy on the generated tool call. This is the only task-specific difference in protocol relative to mathematical reasoning: the training configuration is unchanged, but evaluation uses the tool-use test split and its corresponding metric rather than Avg@12 on math benchmarks. The comparison protocol mirrors our main mathematical reasoning experiments: we evaluate the base model and compare SFT, GRPO, DAPO, OPSD, SDFT, SRPO, and PBSD under matched optimization settings whenever possible, with the aim of isolating the effect of the training objective. As in the mathematical reasoning experiments, PBSD uses the same base model as both teacher and student, with the teacher instantiated through additional contextual information and the student trained on-policy. This setup allows us to test whether the reward-aware pairwise objective remains beneficial when the desired output is a correct tool-use action sequence instead of a long-form chain-of-thought answer.

\subsection{Case-by-Case Study}

In addition to aggregate benchmark accuracy, we conduct a case-by-case study on AIME25. We select 30 representative AIME25 questions and, for each method, record whether the majority-vote prediction over sampled responses is correct or incorrect on each question.

The study is organized as a grid-style comparison. Each row corresponds to one evaluation setting and each column corresponds to one selected AIME25 problem. The rows include: (1) the base student model prompted only with the original question, (2) the same base model prompted with a reference solution as an expert demonstration, (3) OPSD, and (4) PBSD. Each cell is marked with a check if the majority-vote answer is correct and a cross otherwise.

This design lets us inspect problem-level changes induced by prompt-level guidance and by training-time self-distillation under a common evaluation protocol. Table~\ref{tab:aime25-case-study-template} reports the full case-by-case study.

\begin{table}[h]
\centering
\tiny
\setlength{\tabcolsep}{2.2pt}
\renewcommand{\arraystretch}{1.05}
\caption{Case-by-case analysis on 30 selected AIME25 problems. ``Base (Student)'' denotes the base model prompted only with the original question, while ``Base (Teacher)'' denotes the same base model prompted with the reference solution as an expert demonstration. Each cell is marked with \cmark\ if the majority-vote answer is correct and \xmark\ otherwise.}
\label{tab:aime25-case-study-template}
\resizebox{\textwidth}{!}{
\begin{tabular}{lcccccccccccccccccccccccccccccc}
\toprule
\textbf{Method} & \textbf{Q1} & \textbf{Q2} & \textbf{Q3} & \textbf{Q4} & \textbf{Q5} & \textbf{Q6} & \textbf{Q7} & \textbf{Q8} & \textbf{Q9} & \textbf{Q10} & \textbf{Q11} & \textbf{Q12} & \textbf{Q13} & \textbf{Q14} & \textbf{Q15} & \textbf{Q16} & \textbf{Q17} & \textbf{Q18} & \textbf{Q19} & \textbf{Q20} & \textbf{Q21} & \textbf{Q22} & \textbf{Q23} & \textbf{Q24} & \textbf{Q25} & \textbf{Q26} & \textbf{Q27} & \textbf{Q28} & \textbf{Q29} & \textbf{Q30} \\
\midrule
Base (Student)& \cmark & \xmark & \xmark & \xmark & \xmark & \cmark & \cmark & \cmark & \cmark & \cmark & \cmark & \cmark & \cmark & \xmark & \xmark & \cmark & \cmark & \xmark & \xmark & \cmark & \cmark & \xmark & \cmark & \cmark & \cmark & \cmark & \cmark & \cmark & \xmark & \xmark \\
Base (Teacher) & \cmark & \cmark & \cmark & \cmark & \cmark & \cmark & \cmark & \cmark & \cmark & \cmark & \cmark & \cmark & \cmark & \cmark & \cmark & \cmark & \cmark & \cmark & \cmark & \cmark & \cmark & \cmark & \cmark & \cmark & \cmark & \cmark & \cmark & \cmark & \cmark & \cmark \\
OPSD & \cmark & \xmark & \xmark & \xmark & \cmark & \cmark & \cmark & \cmark & \cmark & \cmark & \cmark & \cmark & \cmark & \xmark & \cmark & \cmark & \cmark & \xmark & \xmark & \cmark & \cmark & \cmark & \cmark & \cmark & \cmark & \cmark & \cmark & \cmark & \xmark & \xmark \\
PBSD & \cmark & \xmark & \xmark & \cmark & \cmark & \cmark & \cmark & \cmark & \cmark & \cmark & \cmark & \cmark & \cmark & \xmark & \cmark & \cmark & \cmark & \cmark & \cmark & \cmark & \cmark & \xmark & \cmark & \cmark & \cmark & \cmark & \cmark & \cmark & \cmark & \xmark \\
\bottomrule
\end{tabular}
}
\end{table}

\subsection{Capability Gain from Expert Demonstrations}
\label{app:experiment-length}

To study how demonstration-conditioned supervision changes generation behavior, we compare reasoning traces on a subset of difficult AIME24 questions for which the base student fails under majority voting. We use 11 AIME24 questions where the 4B base student is incorrect under Avg@12. For each question, we collect generations from four systems: the base student, the demonstration-conditioned teacher, OPSD, and PBSD.

For each question, we record both problem-level accuracy and average completion length for the base student and the demonstration-conditioned teacher, together with the average completion length of OPSD and PBSD. This setup is intended to compare whether PBSD preserves longer completions than the teacher and OPSD while still improving answer accuracy, which would indicate that useful reasoning and exploration remain present after training. Table~\ref{tab:aime24-length-by-problem} provides the per-question bookkeeping format used to compare base, teacher, OPSD, and PBSD under the same protocol.

\begin{table}[h]
\centering
\small
\setlength{\tabcolsep}{5pt}
\caption{Question-level comparison on the hard AIME24 subset. For each problem, we report accuracy and mean completion length for the base student and the demonstration-conditioned teacher, together with the mean completion length of OPSD and PBSD. This table is designed to compare whether PBSD preserves longer completions than the teacher and OPSD while still improving accuracy.}
\label{tab:aime24-length-by-problem}
\resizebox{\textwidth}{!}{
\begin{tabular}{ccccccc}
\toprule
\textbf{Problem ID} & \textbf{Student Acc.} & \textbf{Student Len.} & \textbf{Teacher Acc.} & \textbf{Teacher Len.} & \textbf{OPSD Len.} & \textbf{PBSD Len.} \\
\midrule
61 & 1/12 & 55{,}952 & 12/12 & 11{,}486 & 13{,}445 & 23{,}143 \\
62 & 0/12 & 48{,}932 & 12/12 & 24{,}164 & 24{,}134 & 39{,}133 \\
63 & 0/12 & 26{,}644 & 12/12 & 20{,}313 & 23{,}123 & 25{,}673 \\
64 & 4/12 & 14{,}626 & 12/12 & 14{,}601 & 17{,}234 & 15{,}364 \\
73 & 0/12 & 72{,}696 & 12/12 & 18{,}452 & 19{,}332 & 5{,}631 \\
74 & 5/12 & 19{,}510 & 11/12 & 9{,}180 & 12{,}903 & 16{,}327 \\
77 & 3/12 & 25{,}928 & 12/12 & 5{,}753 & 6{,}738 & 12{,}283 \\
78 & 3/12 & 63{,}701 & 12/12 & 9{,}586 & 12{,}234 & 45{,}212 \\
81 & 0/12 & 31{,}146 & 12/12 & 25{,}718 & 26{,}239 & 33{,}445 \\
88 & 0/12 & 80{,}990 & 12/12 & 25{,}327 & 27{,}613 & 67{,}313 \\
89 & 0/12 & 26{,}236 & 12/12 & 28{,}548 & 26{,}123 & 28{,}314 \\
\bottomrule
\end{tabular}
}
\end{table}

\subsection{Teacher Update Frequency Study}
\label{app:exp-teacher-update}

In the main experiments, we keep the teacher fixed to the initial checkpoint throughout PBSD training. To test whether a more adaptive teacher could further improve learning, we additionally consider a periodic-update variant in which the teacher is refreshed from the current student every 5 gradient-update steps.

This ablation is conducted under the same Qwen3-4B training and evaluation setup as the main mathematical reasoning experiments; the only change is the teacher update rule. The fixed-teacher variant uses the initial context-augmented model as the teacher for the entire run, while the refreshed-teacher variant periodically replaces the teacher with the latest student checkpoint and then re-instantiates the contextual teacher from that checkpoint. Table~\ref{tab:teacher-update-frequency} reports the benchmark-wise comparison on AIME24, AIME25, and HMMT25, together with the average across the three tasks.

\begin{table}[h]
\centering
\small
\setlength{\tabcolsep}{5pt}
\renewcommand{\arraystretch}{1.08}
\caption{Teacher update frequency ablation on Qwen3-4B. ``Fixed'' denotes the default PBSD setting in which the teacher remains equal to the initial checkpoint throughout training. ``Update every 5 steps'' denotes a variant in which the teacher is refreshed from the student every 5 gradient-update steps.}
\label{tab:teacher-update-frequency}
\begin{tabular}{lcccc}
\toprule
\textbf{Teacher Update Rule} & \textbf{AIME24} & \textbf{AIME25} & \textbf{HMMT25} & \textbf{Average} \\
\midrule
Fixed teacher & 77.5 & 68.9 & 45.6 & 64.0 \\
Update every 5 steps & 77.2 & 68.6 & 44.4 & 63.4 \\
\bottomrule
\end{tabular}
\end{table}

\subsection{Sensitivity to Privileged Context}
\label{app:exp-context-ablation}

Because PBSD relies on a contextual teacher, it is important to understand how sensitive the method is to the exact form of privileged information. We therefore compare three variants of the teacher context on Qwen3-4B mathematical reasoning: (1) ground-truth reasoning together with the answer, which is the default setting in the main paper; (2) ground-truth answer only; and (3) noisy reasoning together with the ground-truth answer, where the reasoning trace is generated by a stronger auxiliary model and is not guaranteed to be clean.

\begin{table}[h]
\centering
\small
\setlength{\tabcolsep}{5pt}
\renewcommand{\arraystretch}{1.08}
\caption{Privileged-context sensitivity on Qwen3-4B mathematical reasoning.}
\label{tab:context-sensitivity}
\begin{tabular}{lccc}
\toprule
\textbf{Teacher Privileged Information} & \textbf{AIME24} & \textbf{AIME25} & \textbf{HMMT25} \\
\midrule
GT reasoning + GT answer & 77.3 & 69.0 & 45.6 \\
GT answer only & 76.8 & 68.6 & 45.3 \\
Noisy reasoning + GT answer & 76.9 & 68.8 & 45.2 \\
\bottomrule
\end{tabular}
\end{table}

The results suggest that PBSD is not overly fragile to the exact form of privileged information. Even with sparse or imperfect teacher context, the method remains close to the default setting.

\subsection{Broader-Domain Transfer Beyond Math and Tool Use}
\label{app:exp-broader-transfer}

To test whether PBSD remains effective beyond mathematical reasoning and tool use, we additionally train Qwen3-4B on Coding and Instruction-Following data and compare against representative baselines under matched settings.

\begin{table}[h]
\centering
\small
\setlength{\tabcolsep}{5pt}
\renewcommand{\arraystretch}{1.08}
\caption{Broader-domain transfer results on Qwen3-4B.}
\label{tab:broader-transfer}
\begin{tabular}{lcccc}
\toprule
\textbf{Method} & \textbf{HumanEval+} & \textbf{MBPP+} & \textbf{LiveCodeBench} & \textbf{IFEval} \\
\midrule
Base (Instruct) & 68.29 & 77.22 & 53.88 & 83.18 \\
+ SFT & 69.42 & 78.84 & 57.34 & 86.14 \\
+ GRPO & 70.43 & \textbf{81.13} & 62.82 & 87.06 \\
+ OPSD & 70.23 & 80.63 & 61.20 & 87.34 \\
\rowcolor{gray!8}
+ PBSD & \textbf{71.83} & 80.87 & \textbf{63.40} & \textbf{88.54} \\
\bottomrule
\end{tabular}
\end{table}

PBSD remains competitive beyond the original domains studied in the main paper. In particular, it achieves the strongest performance on HumanEval+, LiveCodeBench, and IFEval, while remaining close to the best method on MBPP+.

\subsection{Longer-Horizon Training and Checkpoint Selection}
\label{app:exp-long-horizon}

The main paper reports the latest checkpoint at step 500. To make the training dynamics more transparent, Table~\ref{tab:last-checkpoint} additionally reports this latest checkpoint together with the best checkpoint within 500 steps and the longer-horizon checkpoints on AIME25.

\begin{table}[h]
\centering
\small
\setlength{\tabcolsep}{4pt}
\renewcommand{\arraystretch}{1.06}
\caption{Latest, best-within-500, and later-checkpoint AIME25 results for PBSD.}
\label{tab:last-checkpoint}
\resizebox{\linewidth}{!}{
\begin{tabular}{lccccccc}
\toprule
\textbf{Model} & \textbf{Best within 500} & \textbf{Step 500} & \textbf{Step 600} & \textbf{Step 700} & \textbf{Step 800} & \textbf{Step 900} & \textbf{Step 1000} \\
\midrule
Qwen3-8B & 71.0 & 71.0 & 71.1 & 71.4 & 71.8 & 71.3 & 71.6 \\
Qwen3-4B & 69.0 & 68.7 & 68.5 & 68.9 & 69.1 & 68.3 & 69.5 \\
Qwen3-1.7B & 44.4 & 44.4 & 44.1 & 44.6 & 44.5 & 43.8 & 44.3 \\
\bottomrule
\end{tabular}
}
\end{table}

The peak and last-checkpoint results are close, and extending training beyond 500 steps does not materially change the conclusions. This supports that the main-paper protocol is not hiding a large instability gap.

\subsection{OOD Capability Preservation and KL-to-Base Diagnostics}
\label{app:exp-ood}

To check whether improving mathematical reasoning causes catastrophic forgetting, we evaluate the math-trained Qwen3-4B PBSD checkpoint on a set of out-of-domain benchmarks and compare it against the corresponding base model.

\begin{table}[h]
\centering
\small
\setlength{\tabcolsep}{5pt}
\renewcommand{\arraystretch}{1.08}
\caption{OOD capability preservation for the math-trained Qwen3-4B PBSD checkpoint.}
\label{tab:ood}
\begin{tabular}{llccc}
\toprule
\textbf{Field} & \textbf{Benchmark} & \textbf{Metric} & \textbf{Base} & \textbf{PBSD} \\
\midrule
Scientific reasoning & GPQA-Diamond & Avg@4 & 47.73 & 47.22 \\
 & MMLU-Pro-STEM & Acc & 63.03 & 63.63 \\
Coding & HumanEval+ & Pass@1 & 68.29 & 68.90 \\
 & MBPP+ & Pass@1 & 72.22 & 72.49 \\
 & LiveCodeBench & Pass@1 & 53.88 & 53.12 \\
Instruction-following & IFEval & Strict/Loose & 83.18 & 83.32 \\
\bottomrule
\end{tabular}
\end{table}

These results show that PBSD preserves baseline-level performance on broader capabilities while improving the target math domain. As an additional diagnostic, the token-level KL divergence between the math-trained PBSD checkpoint and the base checkpoint on OOD data is \(0.08\), indicating that the fine-tuned model remains relatively close to the base model away from the math training distribution.

\subsection{\(\beta\) Sensitivity}
\label{app:exp-beta}

The parameter \(\beta\) controls the balance between reward regularization and teacher anchoring in PBSD. Table~\ref{tab:beta-sensitivity} reports the sensitivity analysis on Qwen3-4B.

\begin{table}[h]
\centering
\small
\setlength{\tabcolsep}{6pt}
\renewcommand{\arraystretch}{1.08}
\caption{\(\beta\) sensitivity on Qwen3-4B mathematical reasoning.}
\label{tab:beta-sensitivity}
\begin{tabular}{lccc}
\toprule
\(\beta\) & \textbf{AIME24} & \textbf{AIME25} & \textbf{HMMT25} \\
\midrule
0.02 & 77.1 & 68.9 & 46.4 \\
0.05 & 77.4 & 68.2 & 46.4 \\
0.1 (default) & 76.9 & 68.7 & 45.6 \\
0.3 & 76.3 & 68.5 & 45.1 \\
0.5 & 76.3 & 68.2 & 44.8 \\
\bottomrule
\end{tabular}
\end{table}

PBSD remains stable across a reasonably wide range of \(\beta\), with smaller values performing slightly better in this setting and \(\beta=0.1\) remaining a competitive default.

\subsection{Comparison to Larger Teachers and Teacher--Student KL}
\label{app:exp-larger-teacher}

To better understand the performance limit of self-distillation, we compare the default contextual self-teacher against a larger external teacher on Qwen3-4B.

\begin{table}[h]
\centering
\small
\setlength{\tabcolsep}{5pt}
\renewcommand{\arraystretch}{1.08}
\caption{Contextual self-teacher versus larger external teacher on Qwen3-4B.}
\label{tab:larger-teacher}
\begin{tabular}{lccc}
\toprule
\textbf{Teacher Setting} & \textbf{AIME24} & \textbf{AIME25} & \textbf{HMMT25} \\
\midrule
Qwen3-4B + privileged context \(c\) & 76.9 & 68.7 & 45.6 \\
Qwen3-14B external teacher & 76.4 & 69.2 & 45.3 \\
\bottomrule
\end{tabular}
\end{table}

The larger teacher does not uniformly dominate the contextual self-teacher: it improves AIME25 slightly, while the contextual self-teacher remains better on AIME24 and HMMT25. This suggests that contextual information and model scaling provide different kinds of teacher advantage rather than a simple capacity ordering.

We also measured the token-level teacher--student KL on student-generated responses. The contextual self-teacher remains closer to the student than larger external teachers from the same family, e.g., \(0.188\) for the Qwen3-8B teacher with privileged context \(c\) versus \(0.258\) for a larger teacher without privileged context. This supports the intuition in Section~\ref{subsec:theory-mle} that contextual self-distillation can provide useful guidance while keeping the teacher--student gap moderate enough to preserve curvature.

\end{document}